\documentclass[10pt]{article} 

\usepackage[accepted]{tmlr}
\usepackage[utf8]{inputenc} 
\usepackage[T1]{fontenc}    
\usepackage{url}            
\usepackage{booktabs}       
\usepackage{amsfonts}       
\usepackage{nicefrac}       
\usepackage{microtype}      
\usepackage{algorithm}
\usepackage{tikz}
\usepackage[noend]{algorithmic}
\usepackage{bm}
\usepackage{enumitem}
\usepackage{comment}
\usepackage{dsfont}
\usepackage{amsmath}
\usepackage{amssymb}
\usepackage{amsthm}
\usepackage{mathtools}
\usepackage{thm-restate} 
\usepackage[]{amsbsy}
\usepackage{commath}
\usepackage{physics} 
\usepackage{float} 
\usepackage[hidelinks]{hyperref}
\usepackage{mathrsfs} 
\usepackage[caption=false,font=footnotesize]{subfig}
\usepackage{graphicx}
\usepackage{etoolbox }
\usepackage[capitalize,noabbrev]{cleveref}
\usepackage[skip=0pt]{caption}
\usepackage{xspace} 
\usepackage[normalem]{ulem} 
\usepackage{xcolor}

\let\originalleft\left
\let\originalright\right
\renewcommand{\left}{\mathopen{}\mathclose\bgroup\originalleft}
\renewcommand{\right}{\aftergroup\egroup\originalright}

\newcommand{\bracket}[1]{\left[#1\right]}

\DeclareMathOperator*{\argmax}{argmax} 

\DeclareMathOperator{\E}{\mathbb{E}}

\DeclareMathOperator{\R}{\mathbb{R}}

\renewcommand{\paragraph}[1]{\textbf{#1}\hspace{.1em}}

\theoremstyle{definition}
\newtheorem{definition}{Definition}

\newcommand*{\mat}[1]{\mathbf{#1}}
\newcommand*{\diag}[1]{\textnormal{diag}({#1})}

\makeatletter
\newcommand*\bigcdot{\mathpalette\bigcdot@{1.0}}
\newcommand*\bigcdot@[2]{\mathbin{\vcenter{\hbox{\scalebox{#2}{$\,\m@th#1\bullet\,$}}}}}
\makeatother

\crefname{assumption}{Assumption}{Assumptions}
\crefname{theorem}{Theorem}{Theorems}
\crefname{equation}{}{}
\crefname{ALC@unique}{Line}{Lines}

\newcounter{myalg}
\AtBeginEnvironment{algorithmic}{\refstepcounter{myalg}}
\makeatletter
\newcommand\thefontsize{The current font size is: \f@size pt}
\@addtoreset{ALC@unique}{myalg}
\makeatother

\definecolor{LightGray}{gray}{0.9}

\definecolor{MPL-blue}{HTML}{1F77B4}
\definecolor{MPL-orange}{HTML}{FF7F0E}
\definecolor{MPL-green}{HTML}{2CA02C}
\definecolor{MPL-red}{HTML}{D62728}
\definecolor{MPL-purple}{HTML}{9467BD}
\definecolor{MPL-brown}{HTML}{8C564B}
\definecolor{MPL-pink}{HTML}{E377C2}
\definecolor{MPL-gray}{HTML}{7F7F7F}
\definecolor{MPL-olive}{HTML}{BCBD22}
\definecolor{MPL-cyan}{HTML}{17BECF}

\definecolor{CB-blue}{HTML}{0173B2}
\definecolor{CB-orange}{HTML}{DE8F05}
\definecolor{CB-green}{HTML}{029E73}
\definecolor{CB-red}{HTML}{D55E00}
\definecolor{CB-purple}{HTML}{CC78BC}
\definecolor{CB-brown}{HTML}{CA9161}
\definecolor{CB-pink}{HTML}{FBAFE4}
\definecolor{CB-gray}{HTML}{949494}
\definecolor{CB-olive}{HTML}{ECE133}
\definecolor{CB-cyan}{HTML}{56B4E9}

\newcommand{\ourmethod}{\texttt{vSSM+KF}\xspace}
\newcommand{\vanillassm}{\texttt{vSSM}\xspace}
\newcommand{\noinput}{\texttt{vSSM+KF-u}\xspace}
\newcommand{\mamba}{\texttt{Mamba}\xspace}
\newcommand{\gru}{\texttt{GRU}\xspace}
\newcommand{\transformer}{\texttt{vTransformer}\xspace}
\newcommand{\oracle}{\texttt{Oracle}\xspace}
\newcommand{\nomem}{\texttt{Memoryless}\xspace}

\graphicspath{{figures/}} 

\title{Uncertainty Representations in State-Space Layers for Deep Reinforcement Learning under Partial Observability}

\author{\name Carlos E. Luis \email carlos@robot-learning.de \\
      \addr Bosch Corporate Research \\
      Intelligent Autonomous Systems Group, Technical University Darmstadt
      \AND
      \name Alessandro G. Bottero \email alessandrogiacomo.bottero@bosch.com \\
      \addr Bosch Corporate Research \\
      Intelligent Autonomous Systems Group, Technical University Darmstadt
      \AND
      \name Julia Vinogradska \email julia.vinogradska@bosch.com \\
      \addr Bosch Corporate Research
      \AND
      \name Felix Berkenkamp \email felix.berkenkamp@bosch.com \\
      \addr Bosch Corporate Research
      \AND
      \name Jan Peters \email jan.peters@tu-darmstadt.de \\
      \addr Intelligent Autonomous Systems Group, Technical University Darmstadt \\
      German Research Center for AI (DFKI) \\
      Hessian.AI \\
      Centre for Cognitive Science
}

\def\month{02}  
\def\year{2025} 
\def\openreview{\url{https://openreview.net/forum?id=rfPns0WJyg}} 

\begin{document}

\maketitle

\begin{abstract}
  Optimal decision-making under partial observability requires reasoning about the
  uncertainty of the environment's hidden state. However, most reinforcement learning
  architectures handle partial observability with sequence models that have no internal
  mechanism to incorporate uncertainty in their hidden state representation, such as
  recurrent neural networks, deterministic state-space models and transformers. Inspired
  by advances in probabilistic world models for reinforcement learning, we propose a
  standalone Kalman filter layer that performs closed-form Gaussian inference in linear
  state-space models and train it end-to-end within a model-free architecture to
  maximize returns. Similar to efficient linear recurrent layers, the Kalman filter
  layer processes sequential data using a parallel scan, which scales logarithmically
  with the sequence length. By design, Kalman filter layers are a drop-in replacement
  for other recurrent layers in standard model-free architectures, but importantly they
  include an explicit mechanism for probabilistic filtering of the latent state
  representation. Experiments in a wide variety of tasks with partial observability show
  that Kalman filter layers excel in problems where uncertainty reasoning is key for
  decision-making, outperforming other stateful models.
\end{abstract}

\section{Introduction}
\label{sec:intro}

The classical reinforcement learning (RL) formulation tackles optimal decision-making in
a fully observable Markov decision process (MDP) \citep{sutton_reinforcement_2018}.
However, many real-world problems are \emph{partially} observable, since we only have
access to observations that hide information about the state, e.g., due to noisy
measurements. Learning in partially observable MDPs (POMDPs) is statistically and
computationally intractable in general \citep{papadimitriou_complexity_1987}, but in
many practical scenarios it is theoretically viable \citep{liu_when_2022} and has lead
to successful applications in complex domains like robotics
\citep{zhu_target-driven_2017}, poker \citep{brown_superhuman_2019}, real-time strategy
games \citep{vinyals_grandmaster_2019} and recommendation systems
\citep{li_contextual-bandit_2010}.

Practical algorithms for RL in POMDPs employ sequence models that encode the history of
observations and actions into a latent state representation amenable for policy
optimization. Besides extracting task-relevant information from the history,
probabilistic inference over the latent state is also crucial under partial
observability \citep{kaelbling_planning_1998}. As a motivating example, consider an AI
chatbot that gives restaurant recommendations to users. Since the user's taste (i.e.,
the state) is unknown, the agent must ask questions before ultimately making its
recommendation. Reasoning over the latent state uncertainty is crucial to decide whether
to continue probing the user or end the interaction with a final recommendation. An
optimal agent would gather enough information to recommend a restaurant with a high
likelihood of user satisfaction. In \cref{subsec:best_arm}, we evaluate performance of
our proposed approach in a simplified version of this problem.

A standard recipe for model-free RL in POMDPs is to combine a sequence model (e.g., LSTM
\citep{hochreiter_long_1997}, GRU \citep{cho_properties_2014}) with a policy optimizer
(e.g., PPO \citep{schulman_proximal_2017} or SAC \citep{haarnoja_soft_2018}), which has
shown strong performance in a wide variety of POMDPs \citep{ni_recurrent_2022}. More
recently, transformers \citep{vaswani_attention_2017} have also been adopted as sequence
models in RL showing improved memory capabilities \citep{ni_when_2023}. However, their
inference runtime scales quadratically with the sequence length, which makes them
unsuitable for online learning in physical systems \citep{parisotto_efficient_2020}.
Instead, recent deterministic state-space models (SSMs)
\citep{gu_efficiently_2022,smith_simplified_2023,gu_mamba_2023} maintain the
constant-time inference of stateful models, while achieving logarithmic runtime during
training thanks to efficient parallel scans \citep{smith_simplified_2023}. Moreover,
SSMs have shown improved long-term memory, in-context learning and generalization in RL
\citep{lu_structured_2023}. Yet, in problems where reasoning over latent state
uncertainty is crucial, it remains unclear whether such methods can learn the required
probabilistic inference mechanisms for decision making. The core objective of this
work is to study the role of \emph{explicit} probabilistic inference within a model-free
RL architecture for POMDPs.

While model-free architectures focus on \emph{deterministic} sequence models, in
model-based RL \emph{probabilistic} sequence models are a widespread tool to model
uncertainty in environment dynamics
\citep{watter_embed_2015,hafner_learning_2019,hafner_dream_2020,
becker_uncertainty_2022}. Considering these two sequence modelling approaches, we
concretely investigate the following questions:
\begin{center}
	\textit{Can we leverage the same inference methods developed for model-based RL as general-purpose sequence models in model-free architectures? If so, does it bring any benefits compared to deterministic models?}
\end{center}
Our core hypothesis is that explicit probabilistic inference in sequence models may
serve as an inductive bias to learn in tasks where uncertainty over the latent state is
crucial for decision making, as our motivating example on the restaurant recommendation
chatbot.

\paragraph{Our Contributions.} Inspired by the simple inference scheme in the Recurrent
Kalman Network (RKN) \citep{becker_recurrent_2019} architecture for world models, we
embed closed-form Gaussian inference in linear SSMs as a standalone recurrent layer ---
denoted a Kalman filter (KF) layer --- and train it end-to-end within a model-free
architecture \citep{ni_recurrent_2022} to maximize returns. Since our KF layers are
designed to be a drop-in replacement for standard recurrent layers, they can also be
stacked together and combined with other components (e.g., residual connections,
normalization, etc.) to build more complex sequence models. Similar to
\citet{becker_kalmamba_2024}, we leverage the associative property of the Kalman filter
operations for efficient training of KF layers via parallel scans, which scale
logarithmically with the sequence length provided sufficient parallel GPU cores.

We systematically evaluate our research questions across a variety of POMDPs that
probe distinct capabilities, such as uncertainty reasoning, adaptation,
generalization and filtering of noisy observations. We benchmark the performance of KF
layers against a wide range of baselines from prior work, including GRUs, transformers,
and deterministic SSMs, all embedded in the same model-free architecture. To ensure
fairness in our comparisons, we meticulously control for confounding factors such as
parameter count, training procedure and hyperparameters. By holding all aspects of the
architecture and training constant, we isolate the impact of each sequence model in the
overall performance, providing clear insights into their relative effectiveness. Through
these experiments, we demonstrate that KF layers can be trained effectively end-to-end
on model-free objectives, excelling in tasks where probabilistic inference is key for
decision-making and showing significant improvements over deterministic stateful
models.

\section{Related Work}
\paragraph{RL architectures for POMDPs.} Partial observability in RL tasks requires
agents to maintain \emph{memory} of past interactions. Some approaches incorporate
memory systems inspired by principles of human psychology, such as reward-based learning
\citep{fortunato_generalization_2019}, or rely on mechanisms like context-dependent
retrieval \citep{oh_control_2016}. A more widespread solution involves \emph{sequence
models}, also referred to as \emph{history encoders} \citep{ni_bridging_2024}, which
encode past observations and actions into a state representation useful for RL. These
models have been used to augment policies \citep{wierstra_solving_2007}, value functions
\citep{schmidhuber_networks_1990,bakker_reinforcement_2001} and world models
\citep{schmidhuber_curious_1991,becker_recurrent_2019,shaj_action-conditional_2021,shaj_hidden_2021,shaj_multi_2023}.
This enables RL algorithms, such as DQN \citep{hausknecht_deep_2015}, SAC
\citep{ni_recurrent_2022}, PPO
\citep{kostrikov_pytorch_2018,ni_when_2023,lu_structured_2023}, DPG
\citep{heess_memory-based_2015} and Dyna
\citep{hafner_dream_2020,becker_uncertainty_2022} to handle partial observability. In
this work, we adopt an off-policy model-free architecture similar to
\citet{ni_recurrent_2022}, leveraging its strong performance in various POMDPs.

\paragraph{Sequence models in RL.} Frame-stacking was one of the earliest methods used
in RL to capture temporal context by concatenating consecutive observations
\citep{lin_reinforcement_1993}. It remains a common tool for conveying velocity
information from image-based observations, such as in the Atari benchmark
\citep{bellemare_arcade_2013,mnih_playing_2013}. However, frame-stacking fails to model
long-range dependencies in more complex POMDPs due to its fixed and shallow
representation of temporal relationships. To address this limitation, stateful recurrent
models became the dominant approach for extracting relevant information from arbitrarily
long contexts. Examples include RNNs
\citep{lin_reinforcement_1993,schmidhuber_networks_1990}, LSTMs
\citep{bakker_reinforcement_2001} and GRUs \citep{kostrikov_pytorch_2018}. More
recently, the transformer architecture \citep{vaswani_attention_2017} has shown promise
in improving the long-term memory in RL agents \citep{ni_when_2023}. However, while
transformers excel at modeling long-range dependencies, their slow inference and large
memory footprint reduce their practicality for real-time control tasks, where efficiency
is critical \citep{parisotto_efficient_2020}. These challenges emphasize the need for
more efficient sequence models that balance representational power with computational
feasibility.

\paragraph{Deterministic SSMs.} State-space models are of particular interest to the RL
community due to their computational efficiency compared to traditional sequence models
like RNNs and transformers. They maintain the fast inference of RNNs, but scale
logarithmically (rather than linearly) with the sequence length during training
\citep{smith_simplified_2023}. Moreover, they also circumvent vanishing/exploding
gradients with proper initialization \citep{gu_hippo_2020} and match (or even exceed)
the performance of transformers in long-range sequence modelling tasks
\citep{lu_structured_2023}. In particular, \emph{structured} state space models such as
S4 \citep{gu_efficiently_2022}, S5 \citep{smith_simplified_2023} and S6 / Mamba
\citep{gu_mamba_2023} have emerged as a strong competitor to transformers in general
sequence modelling problems like language \citep{fu_hungry_2023}, audio
\citep{goel_its_2022} and video \citep{nguyen_s4nd_2022}. The adoption of these models
in RL is still in its infancy, however. For instance, \citet{morad_popgym_2023} report
bad performance of a variant of S4 \citep{gu_parameterization_2022} in various POMDPs,
while \citet{lu_structured_2023} show that combining S5 with PPO yields strong results
in long-term memory and in-context learning. These mixed findings suggest that the
performance of deterministic SSMs in RL is sensitive to implementation details and
possibly environtment-dependent. Furthermore, it remains unclear how these models
perform in tasks where uncertainty in the latent state is critical for decision-making,
as they lack explicit probabilistic inference mechanisms. We hypothesize that
probabilistic inference is vital to handle such problems.

\paragraph{Probabilistic SSMs.} Probabilistic SSMs are a common tool in model-based
RL to train both discriminative \citep{haarnoja_backprop_2016} and generative
\citep{ha_recurrent_2018} models of the environment, often referred to as world models.
These models are trained to capture the environment's dynamics, which can then be used
for: (i) planning \citep{hafner_learning_2019} or policy optimization
\citep{becker_uncertainty_2022,hafner_dream_2020} via latent imagination (i.e.,
generating imaginary policy rollouts auto-regressively), or (ii) policy optimization on
the learned latent representation \citep{becker_kalmamba_2024}. A prominent approach is
the Recurrent State Space Model (RSSM) proposed by \citet{hafner_learning_2019}, which
divides the latent state into deterministic and stochastic components and uses a GRU to
propagate the deterministic part forward. More recently, GRUs have been replaced by
transformers \citep{chen_transdreamer_2021} and S4 \citep{samsami_mastering_2024}
models, albeit in a simplified inference scheme that conditions only on the current
observation rather than the history, possibly for computational efficiency. A
common objective function for training these probabilistic SSMs is the evidence lower
bound (ELBO), which provides a lower bound on the log-likelihood of the environment's
data. This ensures that generative models produce plausible trajectories given an action
sequence, typically optimized with variational autoencoders
\citep{kingma_auto-encoding_2014}. These autoencoders shape a low-dimensional latent
representation to capture salient features of the environment's data generation process.
The world model objective can be viewed as an auxiliary loss that helps the sequence
model learn a useful representation for control tasks, which has shown improved sample
efficiency in problems with complex, high-dimensional observations like images
\citep{hafner_mastering_2023}. Other approaches, such as contrastive learning
\citep{laskin_curl_2020}, similarly propose a proxy objective that shapes the learned
representation to maximize agreement between augmented views of the environment. These
auxiliary losses has benefits and drawbacks: they provide a strong learning signal for
representation learning, even in the absence of a reward signal, but they introduce
complexity in training and can interfere with the RL objective, known as the objective
mismatch problem \citep{lambert_objective_2020}. While these tradeoffs warrant research
on their own, the purpose of this work is to bring understanding in the role of
probabilistic inference in model-free RL architectures \emph{without} auxiliary
objectives.

\paragraph{Kalman filters.} A particular class of probabilistic SSMs of relevance
to this work are Kalman filters \citep{kalman_new_1960}, which perform optimal inference
in \emph{linear} SSMs under a Gaussian noise assumption. Since then, Kalman filters have
been theoretically extended to handle non-linear dynamics \citep{serra_kalman_2018} and
also widely adopted in a range of science and engineering fields
\citep{auger_industrial_2013}, including robotics \citep{urrea_kalman_2021}, vision
\citep{chen_kalman_2012}, signal processing and sensor fusion
\citep{khaleghi_multisensor_2013}. They also have a rich history within the machine
learning community, particularly in early applications for time-series forecasting
(Shumway and Stoffer, 1982). While the linear-Gaussian assumption in standard Kalman
filtering is restrictive for the high-dimensional data often found in machine learning
applications \citep{murphy_machine_2012,christopher_m_bishop_pattern_2006}, several
extensions have been proposed. One class of approaches circumvent these limitations by
modelling non-linear state transitions with neural networks
\citep{krishnan_deep_2015,krishnan_structured_2017} and performing approximate inference
via stochastic gradient variational Bayes \citep{kingma_auto-encoding_2014}.
Alternatively, other approaches simplify the inference problem by embedding (locally)
linear-Gaussian SSMs in learned latent spaces
\citep{watter_embed_2015,karl_deep_2017,klushyn_latent_2021}, enabling exact Kalman
filtering and yielding better performance than methods with more complex dynamics but
poor approximate inference \citep{fraccaro_disentangled_2017}. However, the use of full
transition and covariance matrices limits the practical dimensionality of the latent
space and the expressivity of the models. In contrast, we adopt a simpler
parameterization of the linear-Gaussian SSM, using diagonal matrices and covariances,
which significantly reduces the computational burden of Kalman filtering. This approach
scales to higher-dimensional latent spaces, enables logarithmic scaling (in the sequence
length) of the Kalman filter equations \citep{sarkka_temporal_2021,becker_kalmamba_2024}
and preserves the expressivity of the models by offloading representational power to
other components of the architecture, such as encoders and decoders (Haarnoja et al.,
2016; Becker et al., 2019).

\paragraph{Kalman filters in RL.} Kalman filters have been extensively used in RL as
discriminative
\citep{haarnoja_backprop_2016,becker_recurrent_2019,shaj_action-conditional_2021,shaj_hidden_2021,shaj_multi_2023}
or generative \citep{watter_embed_2015,becker_uncertainty_2022,becker_kalmamba_2024}
world models. The former are trained on regression losses to obtain accurate
\emph{predictions}, while the latter are trained with variational inference for
temporally-consistent \emph{generation}. In the model-free architecture considered in
this work, only discriminative sequence models can be integrated without altering the
training procedure; generative models would require auxiliary loss functions, which
would modify the training process and introduce potential confounding factors that are
not part of our experimental design. Closest to our approach is the Recurrent Kalman
Network (RKN) \citep{becker_recurrent_2019}, an encoder-decoder architecture that
employs Kalman filtering using locally linear models and structured (non-diagonal)
covariance matrices. Follow-up work has extended the RKN framework in various ways:
\citet{shaj_action-conditional_2021} include action conditioning,
\citet{shaj_hidden_2021} consider a multi-task setting with hidden task parameters and
\citet{shaj_multi_2023} propose a hierarchical, multi-timescale architecture.  While
these approaches train the latent representation to capture the environment's dynamics,
our work instead focuses on training a similar Kalman filter end-to-end with the RL
objective --- return maximization --- such that the latent space is shaped specifically
for control rather than prediction.

\section{Background}
In this section, we provide the relevant background and introduce core notation used
throughout the paper. We use bold upper case letters ($\mat{A}$) to denote matrices and
calligraphic letters ($\mathcal{X}$) to denote sets. The notation $\diag{\mat{A}}$
refers to a vector containing the diagonal elements for a square matrix $\mat{A}$ and
$\mathcal{P}(\mathcal{X})$ refers to the space of probability distributions over
$\mathcal{X}$.

\subsection{Reinforcement Learning in Partially Observable Markov Decision Processes}
We consider an agent that acts in a finite-horizon partially observable Markov decision
process (POMDP) $\mathcal{M} =\set{\mathcal{S},\mathcal{A}, \mathcal{O},T, p, O, r,
\gamma}$ with state space $\mathcal{S}$, action space $\mathcal{A}$, observation space
$\mathcal{O}$, horizon $T \in \mathbb{N}$, transition function $p: \mathcal{S}
\times \mathcal{A} \to \mathcal{P}(\mathcal{S})$ that maps states and actions to a
probability distribution over $\mathcal{S}$, an emission function $O: \mathcal{S} \to
\mathcal{P}(\mathcal{O})$ that maps states to a probability distribution over
observations, a reward function $r: \mathcal{S} \times \mathcal{A} \to \R$, and a
discount factor $\gamma \in [0,1)$.

At time step $t$ of an episode in $\mathcal{M}$, the agent observes $o_t \sim O(\cdot
\mid s_t)$ and selects an action $a_t \in \mathcal{A}$ based on the observed history
$h_{:t} = (o_{:t}, a_{:t-1}) \in \mathcal{H}_t$, then receives a reward $r_t = r(s_t,
a_t)$ and the next observation $o_{t+1} \sim O(\cdot \mid s_{t+1})$ with $s_{t+1} \sim
p(\cdot \mid s_t, a_t)$.

We adopt the general setting by \citet{ni_when_2023,ni_bridging_2024}, where the RL agent
is equipped with: (\textit{i}) a stochastic policy $\pi: \mathcal{H}_t \to
\mathcal{P}(\mathcal{A})$ that maps from observed history to distribution over actions,
and (\textit{ii}) a value function $Q^\pi: \mathcal{H}_t \times \mathcal{A} \to \R$ that
maps from history and action to the expected return under the policy, defined as
$Q^\pi(h_{:t}, a_t) = \E_\pi \bracket{\sum_{h=t}^T \gamma^{h - t} r_t \mid h_{:t},
a_t}$. The objective of the agent is to find the optimal policy that maximizes the value
starting from some initial state $s_0$, $\pi^\star = \argmax_\pi \E_\pi
\bracket{\sum_{t=0}^{T-1} \gamma^t r_t \mid s_0}$.

\subsection{History Representations}
A weakness of the general formulation of RL in POMDPs is the dependence of both the
policy and the value function on the ever-growing history. Instead, practical algorithms
fight this curse of dimensionality by \emph{compressing} the history into a compact
representation. \citet{ni_bridging_2024} propose to learn such representations via
\emph{history encoders}, defined by a mapping $\phi: \mathcal{H}_t \to \mathcal{Z}$ from
observed history to some latent representation $z_t := \phi(h_{:t}) \in \mathcal{Z}$.
With slight abuse of notation, we denote $\pi(a_t \mid z_t)$ and $Q^\pi(z_t, a_t)$ as
the policy and values under this latent representation, respectively.

\subsection{Probabilistic Inference on Linear SSMs}
We consider time-varying, discrete, linear-Gaussian SSMs defined by
\begin{equation}
  \label{eq:lgssm}
    x_t = \mat{A}_t x_{t-1} + \mat{B}_t u_{t-1} + \varepsilon_t, \quad y_t = \mat{C}_t x_t + \mat{D}_t u_{t-1} +  \nu_t,
\end{equation}
where $t > 0 \in \mathbb{N}$, $x_t \in \R^N$ is the hidden or latent state, $u_t \in
\R^P$ is the input, $y_t \in \R^M$ is the output, $(\mat{A}_t, \mat{B}_t, \mat{C}_t,
\mat{D}_t)$ are matrices of appropriate size, $\varepsilon_t \sim \mathcal{N}(0,
\mat{\Sigma}^{\textnormal{p}}_t)$ and $\nu_t \sim \mathcal{N}(0,
\mat{\Sigma}^{\textnormal{o}}_t)$ are zero-mean process and observation noise variables
with their covariance matrices $\mat{\Sigma}^{\textnormal{p}}_t$ and
$\mat{\Sigma}^{\textnormal{o}}_t$, respectively. Without loss of generality and as it is
common in linear SSMs, we set $\mat{D}_t \equiv \mat{0}$. The latent state
probabilistic model is then $p(x_{t} \mid x_{t - 1}, u_{t-1}) = \mathcal{N}(\mat{A}_t
x_{t-1} + \mat{B}_t u_{t-1}, \mat{\Sigma}^{\textnormal{p}}_t)$ and the observation model
is $p(y_t \mid x_t) = \mathcal{N}(\mat{C}_t x_t, \mat{\Sigma}^{\textnormal{o}}_t)$.
Inference in such a model has a closed-form solution, which is equivalent to the
well-studied Kalman filter \citep{kalman_new_1960}.

\paragraph{Predict.}
The first stage of the Kalman filter propagates forward the \emph{posterior} belief of
the latent state at step $t-1$, given by $\mathcal{N}(x^{+}_{t-1},
\mat{\Sigma}^{+}_{t-1})$, to obtain a \emph{prior} belief at step $t$,
$\mathcal{N}(x^{-}_{t}, \mat{\Sigma}^{-}_{t})$, given by
\begin{equation}
  \label{eq:kf_predict}
  x^{-}_t = \mat{A}_t x^{+}_{t-1} + \mat{B}_t u_{t-1}, \quad 
  \mat{\Sigma}^{-}_{t} = \mat{A}_t\mat{\Sigma}^{+}_{t-1}\mat{A}^{\top}_t + \mat{\Sigma}^{\textnormal{p}}_t.
\end{equation}
\paragraph{Update.} The second stage updates the prior belief at step $t$ given some
observation $w_t$, to obtain the posterior $p(x_t \mid x_{t - 1}, w_t) =
\mathcal{N}(x^{+}_{t}, \Sigma^{+}_{t})$ given by
\begin{equation}
  \label{eq:kf_update}
  x^{+}_t = x^{-}_t + \mat{K}_t(w_t - \mat{C}_t x^{-}_t), \quad \mat{\Sigma}^{+}_{t} = (\mat{I} - \mat{K}_t \mat{C}_t)\mat{\Sigma}_t^{-},
\end{equation}
where $\mat{K}_t = \mat{\Sigma}_t^{-}
\mat{C}_t^{\top}(\mat{C}_t\mat{\Sigma}^{-}_{t}\mat{C}^{\top}_t +
\mat{\Sigma}^{\textnormal{o}}_t)^{-1}$ is known as the Kalman gain. The predict and
update steps are interleaved to process sequences of input and observations $\set{u_t,
w_t}_{t=0}^{K-1}$ of length $K$, starting from some initial belief
$\mathcal{N}(x^{+}_{-1}, \mat{\Sigma}^{+}_{-1})$. 

\subsection{Simplifying Assumptions}
The Kalman filter predict and update equations from \cref{eq:kf_predict,eq:kf_update}
involve expensive matrix multiplication and inversion, which scales poorly with the
latent state dimension $N$. In this section we propose several simplifications, both for easier implementation but also for better scalability.

\paragraph{Time-invariance.} Prior work proposed time-varying SSMs via state-dependent
\citep{becker_recurrent_2019} or input-dependent \citep{gu_mamba_2023} matrices.
Instead, in this work we propose using simple time-invariant matrices, as similarly done
in prior deterministic SSMs \citep{gu_efficiently_2022,smith_simplified_2023}. First,
state-dependent matrices is well motivated by local linearization of dynamics, but they
are incompatible with efficient parallel scan routines, as they break the associative
property of the Kalman filter equations. Second, while input-dependent matrices excel in
associative recall problems where input-selectivity is necessary, Kalman filters provide
similar selection mechanisms via its posterior update (see \cref{subsec:kf_layers}),
without the need for time-varying matrices. Concerning time-invariant process noise,
such simplification reduces the expressivity of the model. However, our initial
experiments with input-dependent process noise showed \emph{worse} performance in RL
than its time-invariant alternative (see \cref{subsec:kf_layers} and
\cref{sec:app_design_ablation}).

\paragraph{Diagonal matrices.} In order to scale to higher-dimensional latent spaces,
prior work in both deterministic and probabilistic SSMs consider \emph{structured} SSMs.
This simply means special structure is imposed into the learnable matrices $(\mat{A},
\mat{B}, \mat{C})$. In particular, we consider a diagonal structure with the HiPPO
initialization proposed in \citet{gu_hippo_2020}, which induces stability in the
recurrence for handling long sequences. In addition, we also consider: (i) diagonal
process and observation noise covariances, (ii) $N=M=P$, which simplifies the
implementation and (iii) identity emission matrices $\mat{C} = \mat{I}$, as proposed in
\citep{becker_uncertainty_2022}. Under this parameterization, the expensive Kalman
filter equations reduce to element-wise operations: 
\begin{equation}
  x^{-}_t = \diag{\mat{A}} \odot x^{+}_{t-1} + \diag{\mat{B}} \odot u_{t-1}, \quad 
  \diag{\mat{\Sigma}^{-}_{t}} = \diag{\mat{A}}^2 \odot \diag{\mat{\Sigma}^{+}_{t-1}} + \diag{\mat{\Sigma}^{\textnormal{p}}_t},
\end{equation}
\begin{equation}
  x^{+}_t = x^{-}_t + \diag{\mat{K}_t} \odot (w_t - x^{-}_t), \quad \diag{\mat{\Sigma}^{+}_{t}} = (\diag{\mat{I}} - \diag{\mat{K}_t}) \odot \diag{\mat{\Sigma}_t^{-}},
\end{equation}
\begin{equation}
  \diag{\mat{K}_t} = \diag{\mat{\Sigma}_t^{-}} \oslash (\diag{\mat{\Sigma}_t^{-}} + \diag{\mat{\Sigma}_t^o}),
\end{equation}
where $\odot$ denoted element-wise vector product and $\oslash$ denotes element-wise
vector division.

\subsection{Parallel Scans}
Efficient implementation of state-space models and Kalman filters employ \emph{parallel
scans} to achieve logarithmic runtime scaling with the sequence length
\citep{smith_simplified_2023,sarkka_temporal_2021}. Given a sequence of elements $(a_0,
a_1, \dots, a_{t-1})$ and an \emph{associative}\footnote{A binary operator $\bigcdot$ is
associative if $(a \bigcdot b) \bigcdot c = a \bigcdot (b \bigcdot c)$ for any triplet
of elements $(a, b, c)$} binary operator $\bigcdot$, the parallel scan algorithm outputs
all the prefix-sums $(a_0, a_0 \bigcdot a_1, \dots, a_0 \bigcdot \dots \bigcdot
a_{t-1})$ in $\mathcal{O}(\log K)$ runtime, given sufficient parallel processors. 

\section{Method: Off-Policy Recurrent Actor-Critic with Kalman filter Layers}
In this section, we describe our method that implements Kalman filtering as a recurrent
layer within a standard actor-critic architecture.

\subsection{General Architecture}
\begin{figure}[t]
  \centering
  \includegraphics[width=0.7\textwidth]{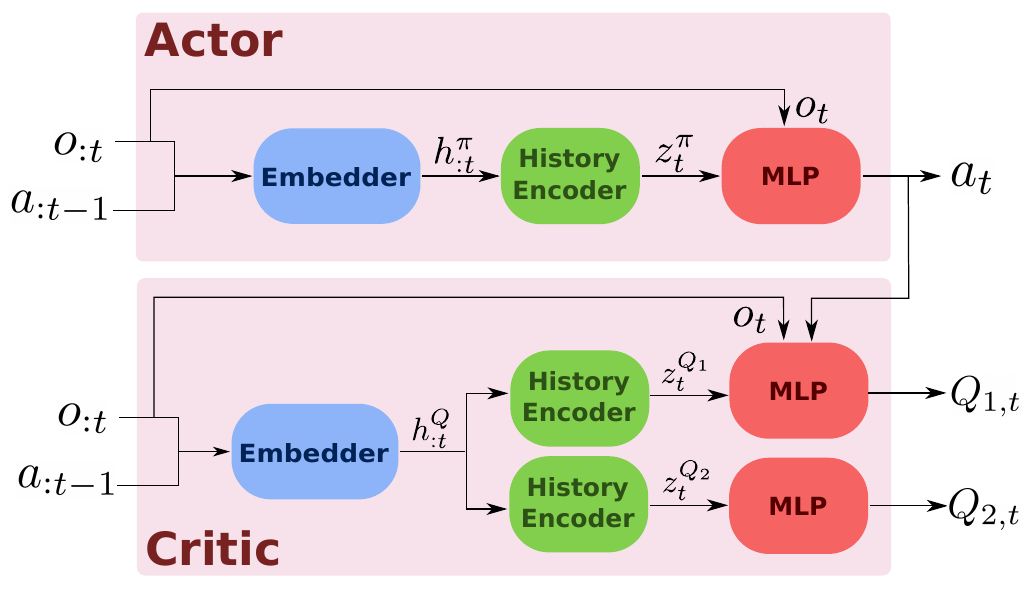}
  \caption{General Recurrent Actor-Critic (RAC) architecture. The components are trained
  end-to-end with the Soft Actor-Critic (SAC) loss function \citep{haarnoja_soft_2018}.
  To handle discrete action spaces, we use the discrete version of SAC by
  \citet{christodoulou_soft_2019}.}
  \label{fig:arch}
\end{figure}

In \cref{fig:arch} we present our Recurrent Actor-Critic (RAC) architecture inspired by
\citet{ni_recurrent_2022}, where we replace the RNN blocks with general history
encoders. We will use this architecture in the following to test the capabilities of different
history encoders in various POMDPs.

For both actor and critic, we embed the sequence of observations and actions into a
single representation $h^{\ast}_{:t}$ which is then passed into the history encoders. We
use a single linear layer as embedder, which we found worked as reliably as more complex
non-linear embedders used in similar RAC architectures by
\citet{morad_popgym_2023,ni_recurrent_2022}. We also include the skip connections from
current observations and actions into the actor-critic heads, as proposed in previous
memory-based architectures \citep{zintgraf_varibad_2021, ni_recurrent_2022}.

\begin{figure}[t]
  \centering
  \includegraphics[width=0.7\textwidth]{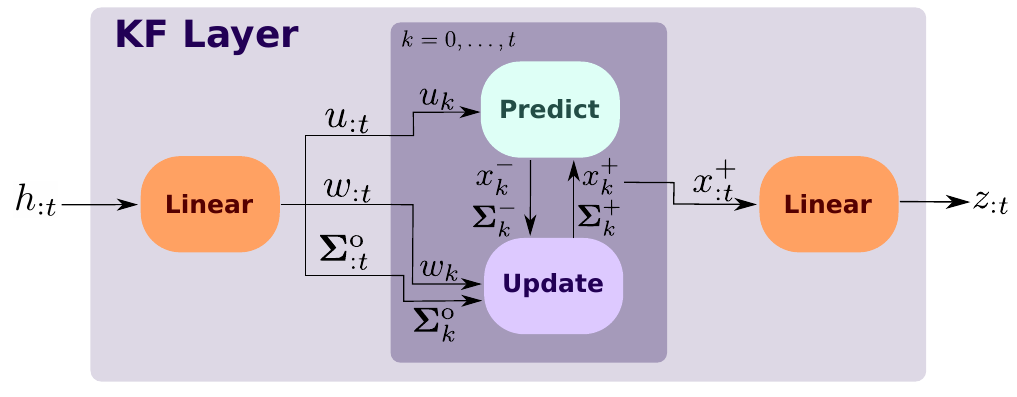}
  \caption{Our proposed Kalman filter layer to build history encoders. The KF layer
  receives a history sequence $h_{:t}$ and projects it into three separate signals in
  latent space: the input $u_{:t}$, the observation $w_{:t}$ and the observation noise
  (diagonal) covariance $\mat{\Sigma}^{\textnormal{o}}_{:t}$. These sequences are
  processed using the standard Kalman filtering equations, which scale logarithmically
  with the sequence length using parallel scans. Lastly, the posterior mean latent state
  $x^{+}_{:t}$ is projected from the latent space back into the history space to obtain
  the compressed representation $z_{:t}$.}
  \label{fig:kf_layers}
\end{figure}

\subsection{Kalman Filter Layers}
\label{subsec:kf_layers}
Our main hypothesis is that principled probabilistic filtering within history encoders
boosts performance in POMDPs, especially those where reasoning about uncertainty is key
for decision-making. To test this hypothesis, we introduce KF layers, as shown in
\cref{fig:kf_layers}. The layer receives as input a history embedding sequence $h_{:t}$
which is then projected into the input $u_{:t}$, observation $w_{:t}$ and observation
noise $\mat{\Sigma}^{\textnormal{o}}_{:t}$ sequences. These three signals serve as input
to the standard KF predict-update equations \cref{eq:kf_predict,eq:kf_update}, which
output a posterior (filtered) latent state $x^{+}_{:t}$. Finally, the posterior sequence
is projected back to the history embedding space to produce the compressed history
representation $z_{:t}$.

\paragraph{History encoders with KF layers.} Similar to recent SSM layers such as S5
\citep{smith_simplified_2023} and S6 \citep{gu_mamba_2023}, these KF layers can be
stacked and combined with other operations such as residual connections, gating
mechanisms, convolutions and normalization to compose a history encoder block in the RAC
architecture. In favor of simplicity, our history encoders are only composed of KF
layers and (optionally) an RMS normalization \citep{zhang_root_2019} output block for
improved stability.

\paragraph{Filtering as a gating mechanism.} We can draw interesting comparisons
between KF layers and other recurrent layers from the perspective of gating mechanisms.
It was shown in Theorem 1 of \citep{gu_mamba_2023} that selective SSMs (S6) behave as
generalized RNN gates through an input-dependent step size $\Delta$. In this case, the
gate depends on the SSM input and controls how much the input influences the next hidden
state. Similarly, as hinted by \citet{becker_recurrent_2019}, during the update step the
Kalman gain is effectively an uncertainty-controlled gate depending on the observation
noise which regulates how much the observation influences the posterior belief over the
latent state. Our experiments in \cref{sec:experiments} shed some light on the strengths
and weaknesses of these approaches for RL under partial observability.

\paragraph{SSM Parameterization.} We follow the procedure in \citet{gu_mamba_2023} and
initialize the continuous-time system $(\mat{\tilde{A}}, \mat{\tilde{B}})$ with HiPPO
matrices. The corresponding discrete-time system $(\mat{A}, \mat{B})$ is obtained via
zero-order hold discretization with a learnable scalar step size $\Delta > 0$
\citep{smith_simplified_2023}.

\paragraph{Design decisions.} We want to highlight two considerations that went
into the design of our KF layers. First, we could generalize the architecture to support
time-varying process noise by including one extra output channel (alongside the input,
observation and observation noise channels) in the history linear projection.
Conceptually, such an input-dependent process noise adds more flexibility to the gating
mechanism implemented within the KF layer, which would be controlled both by the
observation and the process noise signals. Second, we could include the posterior
covariance $\mat{\Sigma}^{+}_{:t}$ as an additional feature for the output linear
projection, alongside the posterior mean $x^{+}_{:t}$. We conduct an ablation study over
these two choices in several continuous control tasks subject to observation noise and
report the results in \cref{sec:app_design_ablation}. The best aggregated performance in
this ablation was obtained with time-invariant process noise and only using the
posterior mean as a feature for the output projection, which empirically justifies our
final design.

\subsection{Masked Associative Operators for Variable Sequence Lengths}
In off-policy RAC architectures, the agent is typically trained with batches of
(sub-)trajectories of possibly different length, sampled from an experience replay
buffer. Thus, history encoders must be able to process batches of variable sequence
length during training. 

A common approach is to right-pad the batch of sequences up to a common length and
ensure the model's output is independent of the padding values. For transformer models,
this can be achieved by using the padding mask as a self-attention mask. For stateful
models like RNNs and SSMs, it is imperative to also output the correct final latent
state for each sequence in the batch. This typically requires a post-processing step
that individually selects for each sequence in the batch the last state before padding.
It turns out that for any recurrent model expressed with an associative operator (e.g.,
SSMs and KFs), we can obtain the correct final state from a batch of padded sequences
\emph{without} additional post-processing by using a parallel scan routine with a
\emph{Masked Associative Operator} (MAO).
\begin{definition}[Masked Associative Operator]
  \label{def:mbo}
  Let $\bigcdot$ be an associative operator acting on elements $e \in \mathcal{E}$, such
  that for any $a, b, c \in \mathcal{E}$, it holds that $(a \bigcdot b) \bigcdot c = a
  \bigcdot (b \bigcdot c)$. Then, the MAO associated with $\bigcdot$, denoted
  $\tilde{\bigcdot}$, acts on elements $\tilde{e} \in \mathcal{E} \times \set{0, 1} =
  (e, m)$, where $m \in \set{0, 1}$ is a binary mask. Then, for $\tilde{a} = (a, m_a)$
  and $\tilde{b} = (b, m_b)$, we have:
  \begin{equation}
    \tilde{a} \tilde{\bigcdot} \tilde{b} = \begin{cases}
      (a \bigcdot b, m_a) \quad &\textnormal{if} \quad m_b = 0 \\
      \tilde{a} \quad &\textnormal{if}  \quad m_b = 1
    \end{cases}
  \end{equation}
\end{definition}
In \cref{app:associative_mao}, we show that any MAO is itself \emph{associative} as long
as we apply a right-padding mask\footnote{A right-padding mask is a sequence $\set{m_0,
m_1, \dots}$ with $m_i \in \set{0, 1}$ such that if $m_i = 1$ then $m_j = 1$ for all $j
> i$.}, thus fulfilling the requirement for parallel scans. In practice, augmenting
existing SSM and KF operators with their MAO counterpart is a minor code change. MAOs
act as a pass-through of the hidden state when padding is applied, thus yielding the
correct state at every time step of the padded sequence for each element of the batch
without additional indexing or bookkeeping. Due to their pass-through nature, MAOs
require strictly equal or less evaluations of the underlying associative operator, which
may yield faster runtimes if the operator is expensive to evaluate and/or many elements
of the input sequence are masked.

\paragraph{MAOs for SSMs and KFs.} As a concrete example, the associative operators for
SSMs and KFs involve matrix product and addition. A compute-efficient implementation of
MAOs for such operators involves sparse matrix operations, where the sparsity is
dictated by the padding mask. However, sparse matrix operations are only expected to
yield better runtime than their dense counterparts for large matrices with sufficient
levels of sparsity, which are not typical in our application. Thus, no speed-up is
expected from using MAOs in the context of this work. 

MAOs are similar to the custom operator proposed by \citet{lu_structured_2023}, but
their effect is fundamentally different: \citet{lu_structured_2023} considers on-policy
RL, where the goal is to handle multi-episode sequences, thus their custom operator
resets the hidden state at episode boundaries. Instead, in our off-policy RAC
architecture, MAOs act as pass-through of the hidden state for padded inputs.

\section{Experiments}
\label{sec:experiments}
In this section, we evaluate the RAC architecture under different history encoders in
various POMDPs. Implementation details and hyperparameters are included in \cref{app:details,app:hparams}, respectively.

\subsection{Baselines}
We consider the following implementation of history encoders within the RAC
architecture.

\paragraph{\vanillassm.} Vanilla, real-valued SSM with diagonal matrices. It is
equivalent to a KF layer with infinite observation noise, i.e., the update step has no
influence on the output. It can also be seen as a simplification of the S4D model
\citep{gu_parameterization_2022}, where states are real-valued rather than complex (as
in Mega \citep{ma_mega_2023}, such that the recurrence can be interpreted as an
exponential moving average) and the recurrence is implemented with a parallel scan
rather than a convolution (as in \citep{smith_simplified_2023}).

\paragraph{\ourmethod.} Probabilistic SSM via the KF layers described in
\cref{fig:kf_layers}. While \vanillassm only \emph{predicts} the next state (i.e.,
the prior in Kalman filtering), \ourmethod additionally \emph{filters} the predicted
state conditioned on the latent observation. Therefore, \ourmethod is equivalent to
\vanillassm with the additional update step of the Kalman filter. Similarly, \vanillassm
is equivalent to \ourmethod with an infinite observation noise variance.

\paragraph{\noinput.} Equivalent to \ourmethod \emph{without} the input signal $u_{:t}$.
It maintains the uncertainty-based gating from the KF layer, but looses flexibility in
the KF predict step to influence the prior belief via the input.

\paragraph{\mamba \citep{gu_mamba_2023}.} Selective state-space model with
input-dependent state transition matrices.

\paragraph{\gru \citep{cho_properties_2014}.} Stateful model with a gating mechanism and
non-linear state transitions.

\paragraph{\transformer \citep{vaswani_attention_2017}.} Vanilla encoder-only
transformer model with sinusoidal positional encoding and causal self-attention.

All SSM-based approaches are implemented using MAOs and parallel scans. Besides these
memory-based agents, we include two additional memoryless agents that implement the same
RAC architecure but without embedders or history encoders.

\paragraph{\oracle.} It has access to the underlying state of the environment,
effectively removing the partial observability aspect of the problem. This method should
upper-bound the performance of history encoders.

\paragraph{\nomem.} Unlike \oracle, it does not have access to the underlying state of
the environment. This method should lower-bound the performance of history encoders.

All the baselines share a common codebase and hyperparameters. For all stateful models,
we use the same latent state dimension $N$ such that parameter count falls within a
$10\%$ tolerance range except for \gru, which naturally has more parameters due to its
gating mechanism (roughly $40 \%$ increase). For \transformer we choose the dimension of
the feed-forward blocks such that the total parameter count is also within $10\%$ of the
SSM methods. With this controlled experimental setup, we aim to evaluate strengths and
weaknesses of the different mechanisms for sequence modelling (gating,
input-selectivity, probabilistic filtering, self-attention) in a wide variety of
partially observable environments.

\begin{figure}[t]
  \centering
  \includegraphics[width=0.9\textwidth]{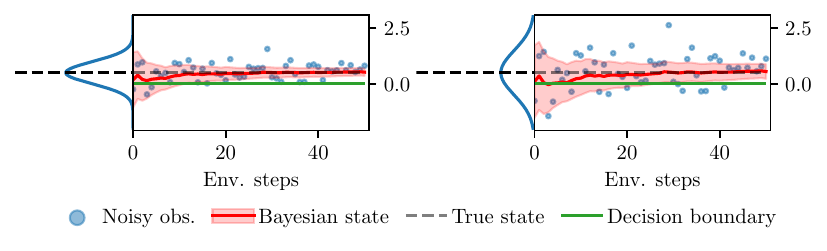}
  \caption{Two example episodes of the Best Arm Identification task of
  \cref{subsec:best_arm}, with $\mu_b = 0.5$ and two different noise scales. \textbf{(Left)}
  Narrow noise distribution with $\sigma_b = 0.5$. \textbf{(Right)} Wide noise
  distribution with $\sigma_b = 1.0$. In red, we visualize the Bayesian posterior mean
  and $3\sigma$ confidence interval around $\mu_b$, obtained via Bayesian linear
  regression using all prior observations in the episode.}
  \label{fig:bestarm_episodes}
\end{figure}

\subsection{Probabilistic Reasoning - Adaptation and Generalization}
\label{subsec:best_arm}
We evaluate probabilistic reasoning capabilities with a carefully designed POMDP that
simplifies our motivating example from \cref{sec:intro}, where an AI chatbot
probes a user in order to recommend a restaurant. Given noisy scalar observations
sampled from a bandit with distribution $\mathcal{N}(\mu_b, \sigma_b)$, the task is to
infer whether the mean $\mu_b$ lies above or below zero. At the start of each episode,
$\mu_b$ and $\sigma_b$ (the latent parameters) are sampled from some given distribution.
Then, at each step of an episode, the RL agent has three choices: (1) request a new
observation from the bandit, which incurs a cost $\rho$, (2) decide the arm has mean
above zero or (3) decide the arm has mean below zero, both of which immediately end the
episode and provide a positive reward if the decision was correct, or a negative reward
if the decision was incorrect. We set a maximum episode length of 1000 steps; if the
agent does not issue a decision by then, it receives the negative reward. Example
rollouts for this environment are provided in \cref{fig:bestarm_episodes}. Given the
Bayesian state from \cref{fig:bestarm_episodes}, an optimal agent must strike a balance
between requesting new information (which reduces uncertainty about the estimated mean)
and minimizing costs. Effective history encoders for this problem should similarly
produce a state representation that encodes uncertainty about the latent parameters.

\begin{figure}[t]
  \centering
  \includegraphics[width=0.9\textwidth]{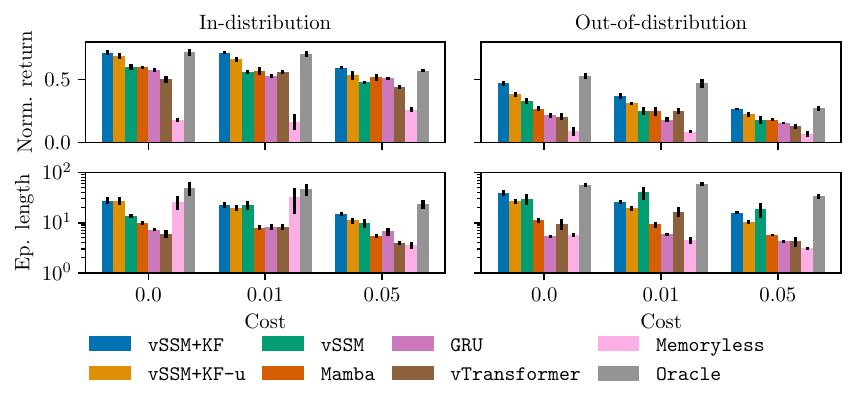}
  \caption{Performance of sequence models in the Best Arm Identification problem after
  $500\textnormal{K}$ environment steps. We conduct experiments for increasing cost of
  requesting new observations and evaluate performance both in and out of distribution,
  averaged over 100 episodes, and report the mean and standard error over 5 random
  seeds. \textbf{(Top row)} Normalized return, obtained by dividing returns by the
  reward given after winning (10 in our case). \textbf{(Bottom row)} Length of
  episodes.}
  \label{fig:best_arm_main_paper}
\end{figure}

We evaluate two core capabilities: adaptation and generalization. Intuitively, an
optimal policy for this problem must be \emph{adaptive} depending on the latent
parameters. For example, if $\mu_b$ is close to zero the agent might need many
observations to make an informed decision, whereas with a large $\abs{\mu_b}$ the
correct decision can be made with few observations. Moreover, we can also evaluate
\emph{generalization} of the learned policy by testing on latent parameters not seen
during training. Our hypothesis is that an agent that learns proper probabilistic
reasoning (e.g., Bayes' rule) should generalize reasonably well in this task.

We conduct experiments for all baselines under increasing cost $\rho$. Instead of
providing the latent parameters directly to the \oracle baseline, we provide the
Bayesian posterior mean and standard deviation around the latent parameter $\mu_b$, as
shown in \cref{fig:bestarm_episodes}. The agents are trained under the latent parameter
distribution given by $\mu_b \sim \textnormal{Unif}(-0.5, 0.5)$ and $\sigma_b \sim
\textnormal{Unif}(0.0, 2.0)$. We additionally evaluate out-of-distribution
generalization by using $\sigma_b^{\text{OOD}} \sim \textnormal{Unif}(2.0, 3.0)$, i.e.,
we test how the agent generalizes to bandits with higher variance. In
\cref{fig:best_arm_main_paper} we report the normalized return and average episode
length for both the training and out-of-distribution latent parameters. Full training
curves are included in \cref{sec:app_bestarm}. \ourmethod achieves the highest return
out of the memory-based agents, both in and out-of-distribution, while matching the
performance of \oracle in-distribution. The better performance of \ourmethod correlates
with longer episodes: compared to the other baselines, \ourmethod learns to request more
observations in order to issue a more informed decision.

\begin{figure}[t]
  \centering
  \includegraphics[width=0.9\textwidth]{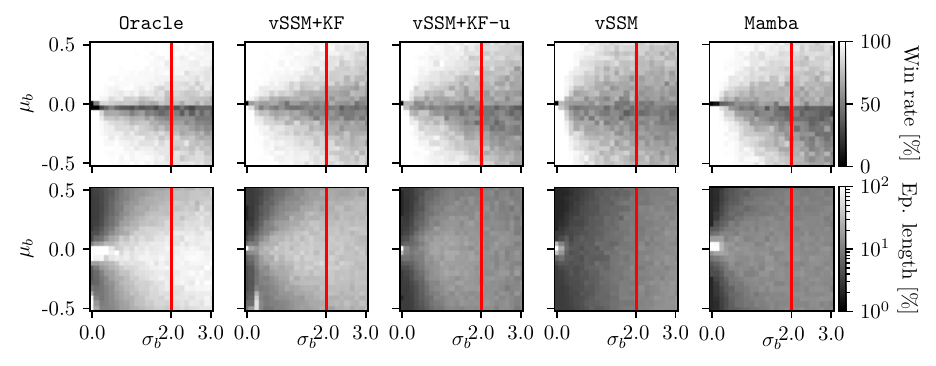}
  \caption{Performance heatmap on Best Arm Identification problem ($\rho=0$). We
  generate a grid of noise parameters $(\mu_b, \sigma_b)$ for a total of 625 unique
  combinations.  The red vertical line separates training (to the left) from
  out-of-distribution (to the right) latent parameters. For each pair of latent
  parameters, we evaluate performance on five independently trained agents over 100
  episodes and report the average win rate and episode lengths.}
  \label{fig:best_arm_offline_eval}
\end{figure}

\paragraph{\ourmethod improves adaptation and generalization.} To gain further insights
on the results, we do a post-training evaluation on a subset of the agents across the
entire latent parameter space, as shown in \cref{fig:best_arm_offline_eval}. \ourmethod
learns adaptation patterns similar to \oracle: the length of episodes increase as the
noise scale $\sigma_b$ increases and decrease as $\abs{\mu_b}$ increases, as it is
intuitively expected. Such adaptation is less pronounced in \vanillassm, \noinput and
\mamba, where episodes are shorter and ultimately results in lower win rates. While
\ourmethod does not match the generalization performance of \oracle, it remains the best
amongst the history encoder baselines. Given our controlled experimental setup, we
attribute the enhanced adaptation and generalization of \ourmethod to the internal
probabilistic filtering implemented in the KF layer. Moreover, comparing \ourmethod and
\noinput highlights that including the input signal in the KF layer leads to improved
performance in this task.

\begin{figure}[t]
  \centering
  \includegraphics[width=\textwidth]{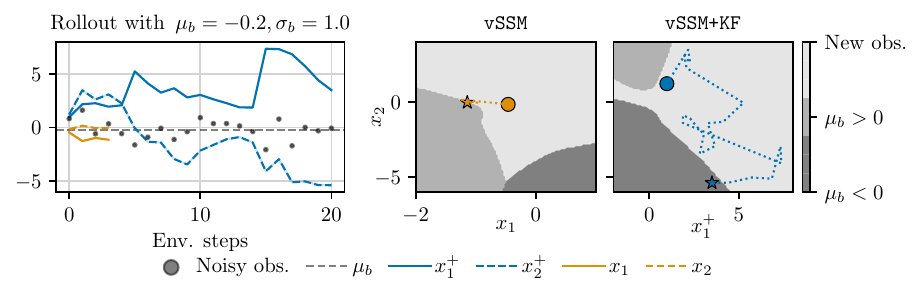}
  \caption{Latent space rollouts in adversarial Best Arm Identification episode.
  \textbf{(Left)} Rollout in latent space ($N=2$) for \ourmethod and \vanillassm after
  training. \textbf{(Middle-Right)} Policy decision boundaries overlaid with the latent
  space trajectory. Circles and stars denote the beginning and end of trajectories,
  respectively.}
  \label{fig:best_arm_rollouts}
\end{figure}

\paragraph{\ourmethod can handle adversarial episodes.} In \cref{fig:best_arm_rollouts}
we compare latent space rollouts\footnote{We use a latent state dimension $N=2$ in order
to plot the policy decision boundary in latent space. This results in slightly worse
performance than the results reported in \cref{fig:best_arm_main_paper}, where we use
$N=128$.} from \ourmethod and \vanillassm in an adversarial episode: $\mu_b$ is
negative, but the first two observations are positive and of relatively large magnitude.
After only four observations, \vanillassm is mislead by the positive observations and
issues the wrong decision, as visualized in \cref{fig:best_arm_rollouts} (middle) where
we show the policy's output across latent space, overlaid with the rollout trajectory.
Instead, \ourmethod remains in the region where the policy requests more observations
before it navigates to the correct region of latent space, as shown in
\cref{fig:best_arm_rollouts} (right). While this example was hand-picked, it is
consistent with the adaptation patterns from \cref{fig:best_arm_offline_eval}.

\subsection{Probabilistic Filtering - Continuous Control under Observation Noise}
\begin{figure}[t]
  \centering
  \includegraphics[width=0.9\textwidth]{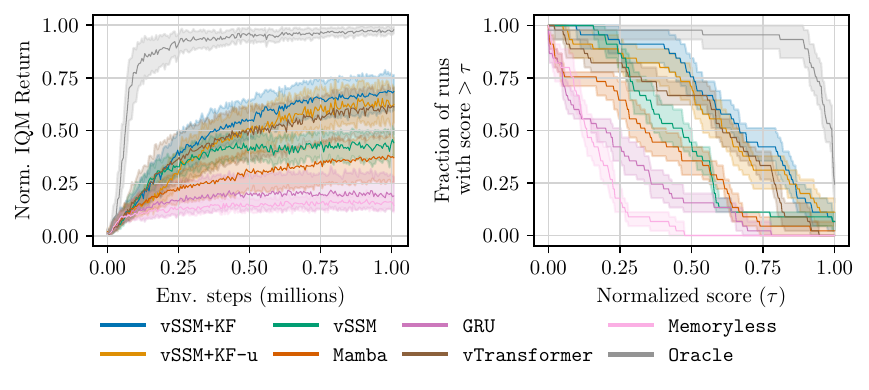}
  \caption{Aggregated performance in noisy DMC benchmark (9 tasks) with 95\% bootstrap
  confidence intervals over five random seeds. \textbf{(Left)} Inter-quartile mean
  returns normalized by the score of \oracle. \textbf{(Right)} Performance profile after
  $1\textnormal{M}$ environment steps. Higher curves correspond to better performance.}
  \label{fig:dmc_rand_main}
\end{figure}

In this experiment, we evaluate the ability to learn control policies subject to
observation noise. Effective history encoders must learn to aggregate observations over
multiple time steps to produce a filtered state representation amenable for control. Our
hypothesis is that internal probabilistic filtering provides an inductive bias for
learning such a filtered representation. To test our hypothesis, we conduct evaluations
across nine environments from the DeepMind Control (DMC) suite
\citep{tunyasuvunakool_dm_control_2020} with zero-mean Gaussian noise added to the
observations, as done by \citet{becker_uncertainty_2022,becker_kalmamba_2024}. We
present aggregated performance in \cref{fig:dmc_rand_main} following the recommendations
from \citep{agarwal_deep_2021}. Detailed training curves are included in
\cref{sec:app_dmc_rand}. We now discuss the main insights from this experiment.

\begin{figure}[t]
  \centering
  \includegraphics[width=1.0\textwidth]{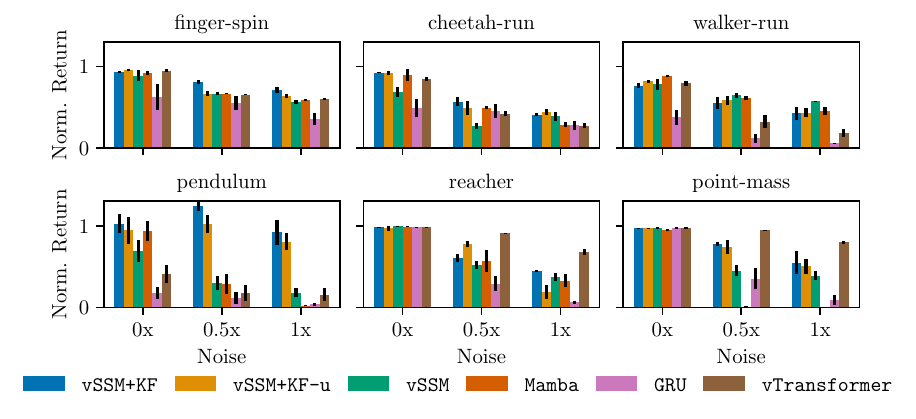}
  \caption{Final performance comparison of recurrent models in six tasks over increasing
  noise levels. We report the mean and standard error over five random seeds (ten for
  \texttt{pendulum} due to large variance) of the return after 1M environment steps,
  normalized by the score of \oracle.}
  \label{fig:dmc_rand_main_noise_ablation}
\end{figure}

\paragraph{\ourmethod improves performance of stateful models.} The KF layer is
the only evaluated add-on for stateful models that significantly improves performance
over the baseline model \vanillassm. This suggests that the uncertainty-based gating in
Kalman filters is more effective at handling noisy data compared to the gating mechanism
implemented by \gru and \mamba. This observation matches the results in the Best Arm
Identification problem from \cref{subsec:best_arm}. Comparing \ourmethod and \noinput,
there is a slight improvement in performance from using an input signal in the KF layer,
but it is not statistically significant.

\paragraph{\ourmethod learns consistently across environments.} From the detailed
results in \cref{fig:dmc_rand_full}, we observe that \ourmethod consistently improves
performance over the \nomem lower-bound and achieves the best or comparable final
performance in five out of nine tasks. Instead, \gru, \mamba and \transformer completely
fail to learn in some tasks, barely matching the performance of \nomem.

We conduct an additional ablation over increasing noise levels in six representative
tasks from the DMC suite, as shown in \cref{fig:dmc_rand_main_noise_ablation}. Training
curves are included in \cref{sec:app_dmc_noise_ablation}

\paragraph{\ourmethod performs close to \oracle under full observability.} We observe
\ourmethod generally matches the performance of \oracle in the absence of noise
(normalized score close to 1.0), whereas \vanillassm and \transformer significantly
underperform in some tasks. This suggests that the added probabilistic filtering in
\ourmethod is a general-purpose strategy even under full observability.

\paragraph{\ourmethod's robustness to noise is environment dependent.}
\cref{fig:dmc_rand_main_noise_ablation} suggests that robustness to noise depends
generally on the environment, without any clear patterns related to task specifics.
\ourmethod is more robust in \texttt{finger-spin}, \texttt{cheetah-run} and
\texttt{pendulum}\footnote{We found that \oracle underperforms in the noiseless
\texttt{pendulum-swingup}, similarly reported in \citep{luis_value-distributional_2023},
which is why the normalized score in this task is larger than 1.0 in some cases.
Moreover, performance does not strictly decrease under higher noise levels, perhaps
because noise may actually help avoid early convergence under sparse rewards.},
\vanillassm is more robust in \texttt{walker-run} but significantly underperforms in
other environments, and \transformer is more robuts in \texttt{reacher} and
\texttt{point-mass} but fails to learn in \texttt{pendulum}. Overall, \ourmethod shows
the most consistent performance across environments and noise levels.

We include additional experiments with noisy DMC tasks in \cref{sec:app_dmc_modelbased},
where we compare performance of \vanillassm and \ourmethod against state-of-the-art
model-based approaches. The main insight from this comparison is that our model-free
approach mostly matches the performance of model-based methods \emph{without} additional
representation learning objectives.

\subsection{General Memory Capabilities}
\begin{figure}[t]
  \centering
  \includegraphics[width=0.9\textwidth]{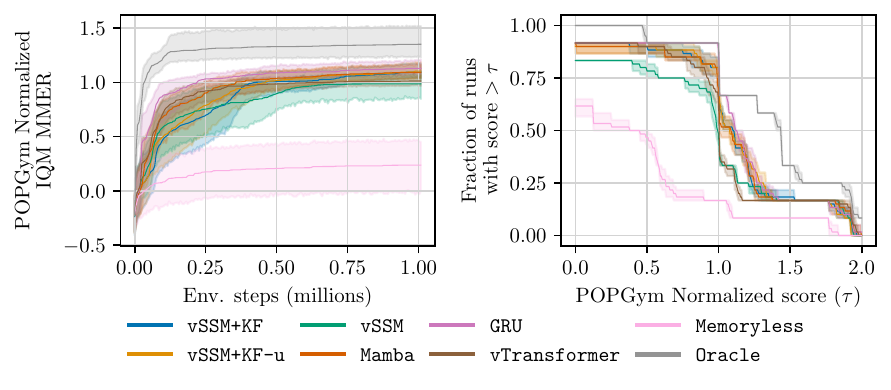}
  \caption{Aggregated performance in POPGym selected environments (12 tasks) with 95\%
  bootstrap confidence intervals over five random seeds. We normalize the maximum-mean
  episodic return (MMER) by the best reported MMER in
  \citep{morad_popgym_2023}\textbf{(Left)} Normalized IQM MMER \textbf{(Right)}
  Performance profile after $1\textnormal{M}$ environment steps. Higher curves
  correspond to better performance and a score of $1.0$ means equivalent performance as
  the best baseline (per environment) reported in POPGym.}
  \label{fig:popgym_main}
\end{figure}

So far the evaluations were conducted in tasks where probabilistic filtering was
intuitively expected to excel. In this experiment, we evaluate performance in a wider
variety of POMDPs from the POPGym \citep{morad_popgym_2023} benchmark. We select a
subset of 12 tasks that probe models for long-term memory, compression, recall, control
under noise and reasoning. The aggregated results are shown in \cref{fig:popgym_main}
and full training curves are also included in \cref{sec:app_popgym}. Below we discuss
the main insights.

\paragraph{KF layers can be generally helpful in POMDPs.} From the performance profile
in \cref{fig:popgym_main} we observe a statistically significant gap between \vanillassm
and \ourmethod. Interestingly, the largest improvements in sample-efficiency
(\texttt{RepeatPreviousEasy}) and final performance (\texttt{MineSweeperEasy})
correspond to tasks that probe for memory duration and recall, respectively. The
parameter count difference between \vanillassm and \ourmethod in these problems is less
than $6\%$, so we believe model capacity is unlikely the reason behind the large
performance difference. We hypothesize that, while probabilistic filtering is not
required to solve these tasks, the KF layer has extra flexibility via the latent
observation and noise signals to accelerate the learning process. We also highlight that
\ourmethod and \noinput show comparable performance in this benchmark, suggesting the
input signal to be less critical in general memory tasks.

\paragraph{\ourmethod is less sample-efficient in pure-memory tasks.} In particular, we
observe that \mamba's input-selectivity is the best-suited mechanism for SSM agents to
solve long-term memory problems, matching the performace of \gru and \transformer. This
is an expected result based on the associative recall performance of Mamba reported in
its original paper \citep{gu_mamba_2023}.

\paragraph{Linear SSMs can have strong performance.} \citet{morad_popgym_2023} report
poor performance when combining PPO with the S4D \citep{gu_parameterization_2022} model.
While we do not evaluate the S4D model and use an off-policy algorithm in our RAC
architecture, our evaluation shows various linear SSMs have strong performance, often
surpassing the best reported scores in \citet{morad_popgym_2023}. Our observation is
consistent with the strong performance of PPO with the S5 model reported by
\citet{lu_structured_2023}.

\subsection{Ablation}
\begin{figure}[t]
  \centering
  \includegraphics[width=0.9\textwidth]{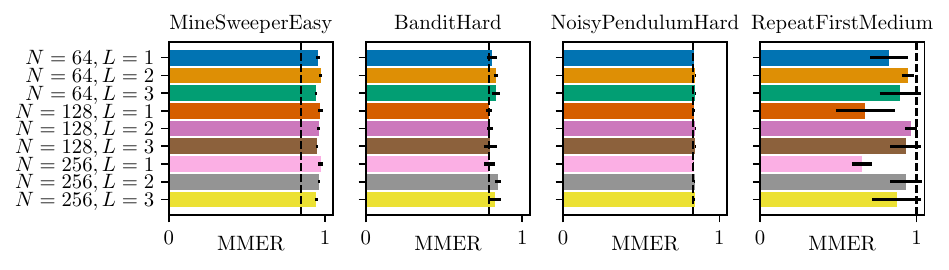}
  \caption{POPGym ablation for \ourmethod over the latent state size $N$ and the number
  of layers $L$. We report the mean and standard error over five random seeds of the
  MMER score after 1M environment steps. The MMER score is shifted from $[-1, 1]$ to
  $[0, 1]$ for easier visualization. The vertical line represents the best score
  reported by \citet{morad_popgym_2023}.}
  \label{fig:popgym_ablation}
\end{figure}

We conduct an ablation on \ourmethod where we vary two hyperparameters: the latent state
size $N$ and the number of stacked KF layers $L$\footnote{We use an RMSNorm output
block in \ourmethod since it was critical to ensure stable learning when $L>1$.}. We
select four representative tasks from POPGym that test different memory capabilities.
The final scores are presented in \cref{fig:popgym_ablation} and the full training
curves are included in \cref{app:popgym_ablation}. Performance is most sensitive to
these hyperparameters in the \texttt{RepeatFirstMedium} task, where the agent must
recall information from the first observation over several steps. The general trend is
that using more than one layer improves final performance and increases
sample-efficiency (see the training curves in \cref{fig:popgym_ablation_full}). Our
results are aligned with the good performance of stacked S5 layers reported by
\citet{lu_structured_2023}, but differ from the observations in \citep{ni_when_2023},
where both LSTM and transformer models performed best with a single layer in a similar
long-term memory task (T-maze passive). From these observations, we believe an
interesting avenue for future work is to study what mechanisms enable effective stacking
and combination of multiple recurrent layers.

\section{Conclusion}
We investigated the use of Kalman filter (KF) layers as sequence models in a recurrent
actor-critic architecture. These layers perform closed-form Gaussian inference in latent
space and output a \emph{filtered} state representation for downstream RL components,
such as value functions and policies. Thanks to the associative nature of the Kalman
filter equations, the KF layers process sequential data efficiently via parallel scans,
whose runtime scales logarithmically with the sequence length. To handle trajectories
with variable length in off-policy RL, we introduced Masked Associative Operators
(MAOs), a general-purpose method that augments any associative operator to recover the
correct hidden state when processing padded input data. The KF layers are used as a
drop-in replacement for RNNs and SSMs in recurrent architectures, and thus can be
trained similarly in an end-to-end, model-free fashion for return maximization.

We evaluated and analysed the strengths and weaknesses of several sequence models in a
wide range of POMDPs. KF layers excel in tasks where uncertainty reasoning is key for
decision-making, such as the Best Arm Identification task and control under observation
noise, significantly improving performance over stateful models like RNNs and
deterministic SSMs. In more general tasks, including long-term memory and associative
recall, KF layers typically match the performance of transformers and other stateful
sequence models, albeit with a lower sample-efficiency.

\paragraph{Limitations and Future Work.} We highlight notable limitations of our
methodology and suggest avenues for future work. First, we investigated two design
decisions in KF layers related to time-varying process noise and posterior covariance as
output features. While they resulted in worse performance (see
\cref{sec:app_design_ablation}), in principle they generalize KF layers and may bring
benefits in other tasks or contexts, so we believe it is worth further investigation.
Second, we use models with relatively low parameter count (< 1M) which is standard in RL
but not on other supervised learning tasks. It may be possible that deeper models with
larger parameter counts enable new capabilities, e.g., probabilistic reasoning, without
explicit probabilistic filtering mechanisms. Third, \ourmethod uses KF layers as
standalone history encoders, but more complex architectures may be needed to stabilize
training at larger parameter counts. Typical strategies found in models like \mamba
include residual connections, layer normalization, convolutions and non-linearities.
Fourth, our evaluations were limited to POMDPS with relatively low-dimensional
observation and action spaces, where small models have enough capacity for learning.
Future work could further evaluate performance in more complex POMDPs (e.g., with
image observations) and compare with our findings.

\subsubsection*{Acknowledgments}
We would like to thank Philipp Becker for providing an efficient parallel scan routine
in Pytorch which we use on all our SSM-based experiments.

\bibliography{references}

\begin{thebibliography}{80}
\providecommand{\natexlab}[1]{#1}
\providecommand{\url}[1]{\texttt{#1}}
\expandafter\ifx\csname urlstyle\endcsname\relax
  \providecommand{\doi}[1]{doi: #1}\else
  \providecommand{\doi}{doi: \begingroup \urlstyle{rm}\Url}\fi

\bibitem[Agarwal et~al.(2021)Agarwal, Schwarzer, Castro, Courville, and Bellemare]{agarwal_deep_2021}
Rishabh Agarwal, Max Schwarzer, Pablo~Samuel Castro, Aaron~C Courville, and Marc Bellemare.
\newblock Deep {Reinforcement} {Learning} at the {Edge} of the {Statistical} {Precipice}.
\newblock In \emph{Advances in {Neural} {Information} {Processing} {Systems}}, volume~34, pages 29304--29320. Curran Associates, Inc., 2021.

\bibitem[Auger et~al.(2013)Auger, Hilairet, Guerrero, Monmasson, Orlowska-Kowalska, and Katsura]{auger_industrial_2013}
François Auger, Mickael Hilairet, Josep~M. Guerrero, Eric Monmasson, Teresa Orlowska-Kowalska, and Seiichiro Katsura.
\newblock Industrial {Applications} of the {Kalman} {Filter}: {A} {Review}.
\newblock \emph{IEEE Transactions on Industrial Electronics}, 60\penalty0 (12):\penalty0 5458--5471, 2013.

\bibitem[Bakker(2001)]{bakker_reinforcement_2001}
Bram Bakker.
\newblock Reinforcement {Learning} with {Long} {Short}-{Term} {Memory}.
\newblock In \emph{Advances in {Neural} {Information} {Processing} {Systems}}, volume~14. MIT Press, 2001.

\bibitem[Becker and Neumann(2022)]{becker_uncertainty_2022}
Philipp Becker and Gerhard Neumann.
\newblock On {Uncertainty} in {Deep} {State} {Space} {Models} for {Model}-{Based} {Reinforcement} {Learning}.
\newblock \emph{Transactions on Machine Learning Research}, July 2022.

\bibitem[Becker et~al.(2019)Becker, Pandya, Gebhardt, Zhao, Taylor, and Neumann]{becker_recurrent_2019}
Philipp Becker, Harit Pandya, Gregor Gebhardt, Cheng Zhao, C.~James Taylor, and Gerhard Neumann.
\newblock Recurrent {Kalman} {Networks}: {Factorized} {Inference} in {High}-{Dimensional} {Deep} {Feature} {Spaces}.
\newblock In \emph{International {Conference} on {Machine} {Learning}}, volume~97, pages 544--552. PMLR, June 2019.

\bibitem[Becker et~al.(2024)Becker, Freymuth, and Neumann]{becker_kalmamba_2024}
Philipp Becker, Niklas Freymuth, and Gerhard Neumann.
\newblock {KalMamba}: {Towards} {Efficient} {Probabilistic} {State} {Space} {Models} for {RL} under {Uncertainty}.
\newblock In \emph{{ICML} {Workshop} on {Aligning} {Reinforcement} {Learning} {Experimentalists} and {Theorists} ({ARLET})}, June 2024.

\bibitem[Bellemare et~al.(2013)Bellemare, Naddaf, Veness, and Bowling]{bellemare_arcade_2013}
Marc~G. Bellemare, Yavar Naddaf, Joel Veness, and Michael Bowling.
\newblock The {Arcade} {Learning} {Environment}: {An} {Evaluation} {Platform} for {General} {Agents}.
\newblock \emph{Journal of Artificial Intelligence Research}, 47:\penalty0 253--279, June 2013.

\bibitem[Bishop(2006)]{christopher_m_bishop_pattern_2006}
Christopher~M. Bishop.
\newblock \emph{Pattern {Recognition} and {Machine} {Learning}}, volume~29.
\newblock Springer, 2006.

\bibitem[Brown and Sandholm(2019)]{brown_superhuman_2019}
Noam Brown and Tuomas Sandholm.
\newblock Superhuman {AI} for {Multiplayer} {Poker}.
\newblock \emph{Science}, 365\penalty0 (6456):\penalty0 885--890, 2019.

\bibitem[Chen et~al.(2021)Chen, Wu, Yoon, and Ahn]{chen_transdreamer_2021}
Chang Chen, Yi-Fu Wu, Jaesik Yoon, and Sungjin Ahn.
\newblock {TransDreamer}: {Reinforcement} {Learning} with {Transformer} {World} {Models}.
\newblock In \emph{{DeepRL} {Workshop} {NeurIPS}}. arXiv, 2021.

\bibitem[Chen(2012)]{chen_kalman_2012}
S.~Y. Chen.
\newblock Kalman {Filter} for {Robot} {Vision}: {A} {Survey}.
\newblock \emph{IEEE Transactions on Industrial Electronics}, 59\penalty0 (11):\penalty0 4409--4420, 2012.

\bibitem[Cho et~al.(2014)Cho, van Merriënboer, Bahdanau, and Bengio]{cho_properties_2014}
Kyunghyun Cho, Bart van Merriënboer, Dzmitry Bahdanau, and Yoshua Bengio.
\newblock On the {Properties} of {Neural} {Machine} {Translation}: {Encoder}–{Decoder} {Approaches}.
\newblock In \emph{8th {Workshop} on {Syntax}, {Semantics} and {Structure} in {Statistical} {Translation}}, pages 103--111. Association for Computational Linguistics (ACL), 2014.

\bibitem[Christodoulou(2019)]{christodoulou_soft_2019}
Petros Christodoulou.
\newblock Soft {Actor}-{Critic} for {Discrete} {Action} {Settings}.
\newblock \emph{arXiv:1910.07207}, October 2019.

\bibitem[D'Oro et~al.(2022)D'Oro, Schwarzer, Nikishin, Bacon, Bellemare, and Courville]{doro_sample-efficient_2022}
Pierluca D'Oro, Max Schwarzer, Evgenii Nikishin, Pierre-Luc Bacon, Marc~G. Bellemare, and Aaron Courville.
\newblock Sample-{Efficient} {Reinforcement} {Learning} by {Breaking} the {Replay} {Ratio} {Barrier}.
\newblock September 2022.

\bibitem[Fortunato et~al.(2019)Fortunato, Tan, Faulkner, Hansen, Puigdomènech~Badia, Buttimore, Deck, Leibo, and Blundell]{fortunato_generalization_2019}
Meire Fortunato, Melissa Tan, Ryan Faulkner, Steven Hansen, Adrià Puigdomènech~Badia, Gavin Buttimore, Charles Deck, Joel~Z Leibo, and Charles Blundell.
\newblock Generalization of {Reinforcement} {Learners} with {Working} and {Episodic} {Memory}.
\newblock In \emph{Advances in {Neural} {Information} {Processing} {Systems}}, volume~32. Curran Associates, Inc., 2019.

\bibitem[Fraccaro et~al.(2017)Fraccaro, Kamronn, Paquet, and Winther]{fraccaro_disentangled_2017}
Marco Fraccaro, Simon Kamronn, Ulrich Paquet, and Ole Winther.
\newblock A {Disentangled} {Recognition} and {Nonlinear} {Dynamics} {Model} for {Unsupervised} {Learning}.
\newblock In \emph{Advances in {Neural} {Information} {Processing} {Systems}}, volume~30. Curran Associates, Inc., 2017.

\bibitem[Fu et~al.(2023)Fu, Dao, Saab, Thomas, Rudra, and Ré]{fu_hungry_2023}
Daniel~Y. Fu, Tri Dao, Khaled~K. Saab, Armin~W. Thomas, Atri Rudra, and Christopher Ré.
\newblock Hungry {Hungry} {Hippos}: {Towards} {Language} {Modeling} with {State} {Space} {Models}.
\newblock In \emph{International {Conference} on {Learning} {Representations}}, 2023.

\bibitem[Goel et~al.(2022)Goel, Gu, Donahue, and Re]{goel_its_2022}
Karan Goel, Albert Gu, Chris Donahue, and Christopher Re.
\newblock It’s {Raw}! {Audio} {Generation} with {State}-{Space} {Models}.
\newblock In \emph{International {Conference} on {Machine} {Learning}}, volume 162, pages 7616--7633. PMLR, July 2022.

\bibitem[Gu and Dao(2023)]{gu_mamba_2023}
Albert Gu and Tri Dao.
\newblock Mamba: {Linear}-{Time} {Sequence} {Modeling} with {Selective} {State} {Spaces}.
\newblock \emph{arXiv:2312.00752}, 2023.

\bibitem[Gu et~al.(2020)Gu, Dao, Ermon, Rudra, and Ré]{gu_hippo_2020}
Albert Gu, Tri Dao, Stefano Ermon, Atri Rudra, and Christopher Ré.
\newblock {HiPPO}: {Recurrent} {Memory} with {Optimal} {Polynomial} {Projections}.
\newblock In \emph{Advances in {Neural} {Information} {Processing} {Systems}}, volume~33, pages 1474--1487. Curran Associates, Inc., 2020.

\bibitem[Gu et~al.(2022{\natexlab{a}})Gu, Goel, and Ré]{gu_efficiently_2022}
Albert Gu, Karan Goel, and Christopher Ré.
\newblock Efficiently {Modeling} {Long} {Sequences} with {Structured} {State} {Spaces}.
\newblock In \emph{International {Conference} on {Learning} {Representations}}, 2022{\natexlab{a}}.

\bibitem[Gu et~al.(2022{\natexlab{b}})Gu, Gupta, Goel, and Ré]{gu_parameterization_2022}
Albert Gu, Ankit Gupta, Karan Goel, and Christopher Ré.
\newblock On the {Parameterization} and {Initialization} of {Diagonal} {State} {Space} {Models}.
\newblock In \emph{Advances in {Neural} {Information} {Processing} {Systems}}, volume~35, 2022{\natexlab{b}}.

\bibitem[Ha and Schmidhuber(2018)]{ha_recurrent_2018}
David Ha and Jürgen Schmidhuber.
\newblock Recurrent {World} {Models} {Facilitate} {Policy} {Evolution}.
\newblock In \emph{Advances in {Neural} {Information} {Processing} {Systems}}, volume~31. Curran Associates, Inc., 2018.

\bibitem[Haarnoja et~al.(2016)Haarnoja, Ajay, Levine, and Abbeel]{haarnoja_backprop_2016}
Tuomas Haarnoja, Anurag Ajay, Sergey Levine, and Pieter Abbeel.
\newblock Backprop {KF}: {Learning} {Discriminative} {Deterministic} {State} {Estimators}.
\newblock In \emph{Advances in {Neural} {Information} {Processing} {Systems}}, pages 4383--4391, 2016. Curran Associates Inc.

\bibitem[Haarnoja et~al.(2018)Haarnoja, Zhou, Abbeel, and Levine]{haarnoja_soft_2018}
Tuomas Haarnoja, Aurick Zhou, Pieter Abbeel, and Sergey Levine.
\newblock Soft {Actor}-{Critic}: {Off}-{Policy} {Maximum} {Entropy} {Deep} {Reinforcement} {Learning} with a {Stochastic} {Actor}.
\newblock In \emph{International {Conference} on {Machine} {Learning}}, volume~80, pages 1861--1870. PMLR, July 2018.

\bibitem[Haarnoja et~al.(2019)Haarnoja, Zhou, Hartikainen, Tucker, Ha, Tan, Kumar, Zhu, Gupta, Abbeel, and Levine]{haarnoja_soft_2019}
Tuomas Haarnoja, Aurick Zhou, Kristian Hartikainen, George Tucker, Sehoon Ha, Jie Tan, Vikash Kumar, Henry Zhu, Abhishek Gupta, Pieter Abbeel, and Sergey Levine.
\newblock Soft {Actor}-{Critic} {Algorithms} and {Applications}.
\newblock \emph{arXiv:1812.05905}, January 2019.

\bibitem[Hafner et~al.(2019)Hafner, Lillicrap, Fischer, Villegas, Ha, Lee, and Davidson]{hafner_learning_2019}
Danijar Hafner, Timothy Lillicrap, Ian Fischer, Ruben Villegas, David Ha, Honglak Lee, and James Davidson.
\newblock Learning {Latent} {Dynamics} for {Planning} from {Pixels}.
\newblock In \emph{International {Conference} on {Machine} {Learning}}, volume~97, pages 2555--2565. PMLR, June 2019.

\bibitem[Hafner et~al.(2020)Hafner, Lillicrap, Ba, and Norouzi]{hafner_dream_2020}
Danijar Hafner, Timothy Lillicrap, Jimmy Ba, and Mohammad Norouzi.
\newblock Dream to {Control}: {Learning} {Behaviors} by {Latent} {Imagination}.
\newblock In \emph{International {Conference} on {Learning} {Representations}}, 2020.

\bibitem[Hafner et~al.(2023)Hafner, Pasukonis, Ba, and Lillicrap]{hafner_mastering_2023}
Danijar Hafner, Jurgis Pasukonis, Jimmy Ba, and Timothy Lillicrap.
\newblock Mastering {Diverse} {Domains} through {World} {Models}, January 2023.
\newblock URL \url{http://arxiv.org/abs/2301.04104}.

\bibitem[Hausknecht and Stone(2015)]{hausknecht_deep_2015}
Matthew Hausknecht and Peter Stone.
\newblock Deep {Recurrent} {Q}-{Learning} for {Partially} {Observable} {MDPs}.
\newblock In \emph{{AAAI} {Fall} {Symposium} {Series}}, 2015.

\bibitem[Heess et~al.(2015)Heess, Hunt, Lillicrap, and Silver]{heess_memory-based_2015}
Nicolas Heess, Jonathan~J. Hunt, Timothy~P. Lillicrap, and David Silver.
\newblock Memory-{Based} {Control} with {Recurrent} {Neural} {Networks}.
\newblock In \emph{{NIPS} {Deep} {Reinforcement} {Learning} {Workshop}}, December 2015.

\bibitem[Hochreiter and Schmidhuber(1997)]{hochreiter_long_1997}
Sepp Hochreiter and Jürgen Schmidhuber.
\newblock Long {Short}-{Term} {Memory}.
\newblock \emph{Neural Computation}, 9\penalty0 (8):\penalty0 1735--1780, 1997.

\bibitem[Kaelbling et~al.(1998)Kaelbling, Littman, and Cassandra]{kaelbling_planning_1998}
Leslie~Pack Kaelbling, Michael~L. Littman, and Anthony~R. Cassandra.
\newblock Planning and {Acting} in {Partially} {Observable} {Stochastic} {Domains}.
\newblock \emph{Artificial Intelligence}, 101\penalty0 (1):\penalty0 99--134, 1998.

\bibitem[Kalman(1960)]{kalman_new_1960}
R.~E. Kalman.
\newblock A {New} {Approach} to {Linear} {Filtering} and {Prediction} {Problems}.
\newblock \emph{Journal of Basic Engineering}, 82\penalty0 (1):\penalty0 35--45, March 1960.

\bibitem[Karl et~al.(2017)Karl, Soelch, Bayer, and Smagt]{karl_deep_2017}
Maximilian Karl, Maximilian Soelch, Justin Bayer, and Patrick van~der Smagt.
\newblock Deep {Variational} {Bayes} {Filters}: {Unsupervised} {Learning} of {State} {Space} {Models} from {Raw} {Data}.
\newblock February 2017.

\bibitem[Khaleghi et~al.(2013)Khaleghi, Khamis, Karray, and Razavi]{khaleghi_multisensor_2013}
Bahador Khaleghi, Alaa Khamis, Fakhreddine~O. Karray, and Saiedeh~N. Razavi.
\newblock Multisensor {Data} {Fusion}: {A} {Review} of the {State}-of-the-art.
\newblock \emph{Information Fusion}, 14\penalty0 (1):\penalty0 28--44, 2013.

\bibitem[Kingma and Welling(2014)]{kingma_auto-encoding_2014}
Diederik~P. Kingma and Max Welling.
\newblock Auto-{Encoding} {Variational} {Bayes}.
\newblock In \emph{International {Conference} on {Learning} {Representations}}, 2014.

\bibitem[Klushyn et~al.(2021)Klushyn, Kurle, Soelch, Cseke, and van~der Smagt]{klushyn_latent_2021}
Alexej Klushyn, Richard Kurle, Maximilian Soelch, Botond Cseke, and Patrick van~der Smagt.
\newblock Latent {Matters}: {Learning} {Deep} {State}-{Space} {Models}.
\newblock In \emph{Advances in {Neural} {Information} {Processing} {Systems}}, volume~34, pages 10234--10245. Curran Associates, Inc., 2021.

\bibitem[Kostrikov(2018)]{kostrikov_pytorch_2018}
Ilya Kostrikov.
\newblock {PyTorch} {Implementations} of {Reinforcement} {Learning} {Algorithms}, 2018.
\newblock URL \url{https://github.com/ikostrikov/pytorch-a2c-ppo-acktr-gail}.

\bibitem[Krishnan et~al.(2017)Krishnan, Shalit, and Sontag]{krishnan_structured_2017}
Rahul Krishnan, Uri Shalit, and David Sontag.
\newblock Structured {Inference} {Networks} for {Nonlinear} {State} {Space} {Models}.
\newblock \emph{Proceedings of the AAAI Conference on Artificial Intelligence}, 31\penalty0 (1), February 2017.

\bibitem[Krishnan et~al.(2015)Krishnan, Shalit, and Sontag]{krishnan_deep_2015}
Rahul~G. Krishnan, Uri Shalit, and David Sontag.
\newblock Deep {Kalman} {Filters}, November 2015.
\newblock URL \url{http://arxiv.org/abs/1511.05121}.

\bibitem[Lambert et~al.(2020)Lambert, Amos, Yadan, and Calandra]{lambert_objective_2020}
Nathan Lambert, Brandon Amos, Omry Yadan, and Roberto Calandra.
\newblock Objective {Mismatch} in {Model}-based {Reinforcement} {Learning}.
\newblock In \emph{Learning for {Dynamics} and {Control}}, pages 761--770. PMLR, 2020.

\bibitem[Laskin et~al.(2020)Laskin, Srinivas, and Abbeel]{laskin_curl_2020}
Michael Laskin, Aravind Srinivas, and Pieter Abbeel.
\newblock {CURL}: {Contrastive} {Unsupervised} {Representations} for {Reinforcement} {Learning}.
\newblock In \emph{International {Conference} on {Machine} {Learning}}, volume 119, pages 5639--5650. PMLR, July 2020.

\bibitem[Li et~al.(2010)Li, Chu, Langford, and Schapire]{li_contextual-bandit_2010}
Lihong Li, Wei Chu, John Langford, and Robert~E. Schapire.
\newblock A {Contextual}-{Bandit} {Approach} to {Personalized} {News} {Article} {Recommendation}.
\newblock In \emph{International {Conference} on {World} {Wide} {Web}}, pages 661--670, 2010. Association for Computing Machinery.

\bibitem[Lin and Mitchell(1993)]{lin_reinforcement_1993}
Long-Ji Lin and Tom~M Mitchell.
\newblock Reinforcement {Learning} with {Hidden} {States}.
\newblock In \emph{International {Conference} on {Simulation} of {Adaptive} {Behavior}}, pages 271--280. MIT Press, 1993.

\bibitem[Liu et~al.(2022)Liu, Chung, Szepesvari, and Jin]{liu_when_2022}
Qinghua Liu, Alan Chung, Csaba Szepesvari, and Chi Jin.
\newblock When {Is} {Partially} {Observable} {Reinforcement} {Learning} {Not} {Scary}?
\newblock In \emph{Conference on {Learning} {Theory}}, pages 5175--5220. PMLR, June 2022.

\bibitem[Lu et~al.(2023)Lu, Schroecker, Gu, Parisotto, Foerster, Singh, and Behbahani]{lu_structured_2023}
Chris Lu, Yannick Schroecker, Albert Gu, Emilio Parisotto, Jakob Foerster, Satinder Singh, and Feryal Behbahani.
\newblock Structured {State} {Space} {Models} for {In}-{Context} {Reinforcement} {Learning}.
\newblock In \emph{Advances in {Neural} {Information} {Processing} {Systems}}, volume~36, pages 47016--47031. Curran Associates, Inc., 2023.

\bibitem[Luis et~al.(2023)Luis, Bottero, Vinogradska, Berkenkamp, and Peters]{luis_value-distributional_2023}
Carlos~E. Luis, Alessandro~G. Bottero, Julia Vinogradska, Felix Berkenkamp, and Jan Peters.
\newblock Value-{Distributional} {Model}-{Based} {Reinforcement} {Learning}.
\newblock \emph{arXiv:2308.06590}, August 2023.

\bibitem[Ma et~al.(2023)Ma, Zhou, Kong, He, Gui, Neubig, May, and Zettlemoyer]{ma_mega_2023}
Xuezhe Ma, Chunting Zhou, Xiang Kong, Junxian He, Liangke Gui, Graham Neubig, Jonathan May, and Luke Zettlemoyer.
\newblock Mega: {Moving} {Average} {Equipped} {Gated} {Attention}.
\newblock In \emph{International {Conference} on {Learning} {Representations}}, 2023.

\bibitem[Mnih et~al.(2013)Mnih, Kavukcuoglu, Silver, Graves, Antonoglou, Wierstra, and Riedmiller]{mnih_playing_2013}
Volodymyr Mnih, Koray Kavukcuoglu, David Silver, Alex Graves, Ioannis Antonoglou, Daan Wierstra, and Martin Riedmiller.
\newblock Playing {Atari} with {Deep} {Reinforcement} {Learning}.
\newblock In \emph{{NIPS} {Deep} {Learning} {Workshop}}, December 2013.

\bibitem[Morad et~al.(2023)Morad, Kortvelesy, Bettini, Liwicki, and Prorok]{morad_popgym_2023}
Steven Morad, Ryan Kortvelesy, Matteo Bettini, Stephan Liwicki, and Amanda Prorok.
\newblock {POPGym}: {Benchmarking} {Partially} {Observable} {Reinforcement} {Learning}.
\newblock In \emph{International {Conference} on {Learning} {Representations}}, 2023.

\bibitem[Murphy(2012)]{murphy_machine_2012}
Kevin Murphy.
\newblock \emph{Machine {Learning}: {A} {Probabilistic} {Perspective}}.
\newblock MIT Press, 2012.

\bibitem[Nguyen et~al.(2022)Nguyen, Goel, Gu, Downs, Shah, Dao, Baccus, and Ré]{nguyen_s4nd_2022}
Eric Nguyen, Karan Goel, Albert Gu, Gordon Downs, Preey Shah, Tri Dao, Stephen Baccus, and Christopher Ré.
\newblock {S4ND}: {Modeling} {Images} and {Videos} as {Multidimensional} {Signals} with {State} {Spaces}.
\newblock In \emph{Advances in {Neural} {Information} {Processing} {Systems}}, volume~35, pages 2846--2861. Curran Associates, Inc., 2022.

\bibitem[Ni et~al.(2022)Ni, Eysenbach, and Salakhutdinov]{ni_recurrent_2022}
Tianwei Ni, Benjamin Eysenbach, and Ruslan Salakhutdinov.
\newblock Recurrent {Model}-{Free} {RL} {Can} {Be} a {Strong} {Baseline} for {Many} {POMDPs}.
\newblock In \emph{International {Conference} on {Machine} {Learning}}, pages 16691--16723. PMLR, June 2022.

\bibitem[Ni et~al.(2023)Ni, Ma, Eysenbach, and Bacon]{ni_when_2023}
Tianwei Ni, Michel Ma, Benjamin Eysenbach, and Pierre-Luc Bacon.
\newblock When {Do} {Transformers} {Shine} in {RL}? {Decoupling} {Memory} from {Credit} {Assignment}.
\newblock In \emph{Advances in {Neural} {Information} {Processing} {Systems}}, volume~36, pages 50429--50452. Curran Associates, Inc., 2023.

\bibitem[Ni et~al.(2024)Ni, Eysenbach, Seyedsalehi, Ma, Gehring, Mahajan, and Bacon]{ni_bridging_2024}
Tianwei Ni, Benjamin Eysenbach, Erfan Seyedsalehi, Michel Ma, Clement Gehring, Aditya Mahajan, and Pierre-Luc Bacon.
\newblock Bridging {State} and {History} {Representations}: {Understanding} {Self}-{Predictive} {RL}.
\newblock In \emph{International {Conference} on {Learning} {Representations}}, 2024.

\bibitem[Oh et~al.(2016)Oh, Chockalingam, Satinder, and Lee]{oh_control_2016}
Junhyuk Oh, Valliappa Chockalingam, Satinder, and Honglak Lee.
\newblock Control of {Memory}, {Active} {Perception}, and {Action} in {Minecraft}.
\newblock In \emph{International {Conference} on {Machine} {Learning}}, volume~48, pages 2790--2799, June 2016. PMLR.

\bibitem[Papadimitriou and Tsitsiklis(1987)]{papadimitriou_complexity_1987}
Christos~H. Papadimitriou and John~N. Tsitsiklis.
\newblock The {Complexity} of {Markov} {Decision} {Processes}.
\newblock \emph{Mathematics of Operations Research}, 12\penalty0 (3):\penalty0 441--450, 1987.

\bibitem[Parisotto and Salakhutdinov(2020)]{parisotto_efficient_2020}
Emilio Parisotto and Russ Salakhutdinov.
\newblock Efficient {Transformers} in {Reinforcement} {Learning} using {Actor}-{Learner} {Distillation}.
\newblock In \emph{International {Conference} on {Learning} {Representations}}, October 2020.

\bibitem[Paszke et~al.(2019)Paszke, Gross, Massa, Lerer, Bradbury, Chanan, Killeen, Lin, Gimelshein, Antiga, Desmaison, Kopf, Yang, DeVito, Raison, Tejani, Chilamkurthy, Steiner, Fang, Bai, and Chintala]{paszke_pytorch_2019}
Adam Paszke, Sam Gross, Francisco Massa, Adam Lerer, James Bradbury, Gregory Chanan, Trevor Killeen, Zeming Lin, Natalia Gimelshein, Luca Antiga, Alban Desmaison, Andreas Kopf, Edward Yang, Zachary DeVito, Martin Raison, Alykhan Tejani, Sasank Chilamkurthy, Benoit Steiner, Lu~Fang, Junjie Bai, and Soumith Chintala.
\newblock {PyTorch}: {An} {Imperative} {Style}, {High}-{Performance} {Deep} {Learning} {Library}.
\newblock In \emph{Advances in {Neural} {Information} {Processing} {Systems}}, volume~32. Curran Associates, Inc., 2019.

\bibitem[Samsami et~al.(2024)Samsami, Zholus, Rajendran, and Chandar]{samsami_mastering_2024}
Mohammad~Reza Samsami, Artem Zholus, Janarthanan Rajendran, and Sarath Chandar.
\newblock Mastering {Memory} {Tasks} with {World} {Models}.
\newblock In \emph{International {Conference} on {Learning} {Representations}}, 2024.

\bibitem[Sarkka and Garcia-Fernandez(2021)]{sarkka_temporal_2021}
Simo Sarkka and Angel~F. Garcia-Fernandez.
\newblock Temporal {Parallelization} of {Bayesian} {Smoothers}.
\newblock \emph{IEEE Transactions on Automatic Control}, 66\penalty0 (1):\penalty0 299--306, January 2021.

\bibitem[Schmidhuber(1991)]{schmidhuber_curious_1991}
J.~Schmidhuber.
\newblock Curious {Model}-{Building} {Control} {Systems}.
\newblock In \emph{{IEEE} {International} {Joint} {Conference} on {Neural} {Networks}}, pages 1458--1463 vol.2, 1991. IEEE.

\bibitem[Schmidhuber(1990)]{schmidhuber_networks_1990}
Jurgen Schmidhuber.
\newblock Networks {Adjusting} {Networks}.
\newblock In \emph{Distributed {Adaptive} {Neural} {Information} {Processing}}, pages 197--208, 1990.

\bibitem[Schulman et~al.(2017)Schulman, Wolski, Dhariwal, Radford, and Klimov]{schulman_proximal_2017}
John Schulman, Filip Wolski, Prafulla Dhariwal, Alec Radford, and Oleg Klimov.
\newblock Proximal {Policy} {Optimization} {Algorithms}.
\newblock \emph{arXiv:1707.06347}, August 2017.

\bibitem[Serra(2018)]{serra_kalman_2018}
Ginalber Luiz de~Oliveira Serra.
\newblock \emph{Kalman {Filters} - {Theory} for {Advanced} {Applications}}.
\newblock February 2018.

\bibitem[Shaj et~al.(2021{\natexlab{a}})Shaj, Becker, Büchler, Pandya, Duijkeren, Taylor, Hanheide, and Neumann]{shaj_action-conditional_2021}
Vaisakh Shaj, Philipp Becker, Dieter Büchler, Harit Pandya, Niels~van Duijkeren, C.~James Taylor, Marc Hanheide, and Gerhard Neumann.
\newblock Action-{Conditional} {Recurrent} {Kalman} {Networks} {For} {Forward} and {Inverse} {Dynamics} {Learning}.
\newblock In \emph{Conference on {Robot} {Learning}}, volume 155, pages 765--781. PMLR, November 2021{\natexlab{a}}.

\bibitem[Shaj et~al.(2021{\natexlab{b}})Shaj, Büchler, Sonker, Becker, and Neumann]{shaj_hidden_2021}
Vaisakh Shaj, Dieter Büchler, Rohit Sonker, Philipp Becker, and Gerhard Neumann.
\newblock Hidden {Parameter} {Recurrent} {State} {Space} {Models} {For} {Changing} {Dynamics} {Scenarios}.
\newblock In \emph{International {Conference} on {Learning} {Representations}}, October 2021{\natexlab{b}}.

\bibitem[Shaj et~al.(2023)Shaj, Zadeh, Demir, Douat, and Neumann]{shaj_multi_2023}
Vaisakh Shaj, Saleh~Gholam Zadeh, Ozan Demir, Luiz~Ricardo Douat, and Gerhard Neumann.
\newblock Multi {Time} {Scale} {World} {Models}.
\newblock In \emph{Advances in {Neural} {Information} {Processing} {Systems}}, November 2023.

\bibitem[Smith et~al.(2023)Smith, Warrington, and Linderman]{smith_simplified_2023}
Jimmy T.~H. Smith, Andrew Warrington, and Scott Linderman.
\newblock Simplified {State} {Space} {Layers} for {Sequence} {Modeling}.
\newblock In \emph{International {Conference} on {Learning} {Representations}}, 2023.

\bibitem[Sutton and Barto(2018)]{sutton_reinforcement_2018}
Richard Sutton and Andrew Barto.
\newblock \emph{Reinforcement {Learning}: {An} {Introduction}}, volume~7.
\newblock MIT Press, 2018.

\bibitem[Tunyasuvunakool et~al.(2020)Tunyasuvunakool, Muldal, Doron, Liu, Bohez, Merel, Erez, Lillicrap, Heess, and Tassa]{tunyasuvunakool_dm_control_2020}
Saran Tunyasuvunakool, Alistair Muldal, Yotam Doron, Siqi Liu, Steven Bohez, Josh Merel, Tom Erez, Timothy Lillicrap, Nicolas Heess, and Yuval Tassa.
\newblock dm\_control: {Software} and {Tasks} for {Continuous} {Control}.
\newblock \emph{Software Impacts}, 6:\penalty0 100022, 2020.

\bibitem[Urrea and Agramonte(2021)]{urrea_kalman_2021}
Claudio Urrea and Rayko Agramonte.
\newblock Kalman {Filter}: {Historical} {Overview} and {Review} of {Its} {Use} in {Robotics} 60 {Years} {After} {Its} {Creation}.
\newblock \emph{Journal of Sensors}, 2021\penalty0 (1), 2021.

\bibitem[Vaswani et~al.(2017)Vaswani, Shazeer, Parmar, Uszkoreit, Jones, Gomez, Kaiser, and Polosukhin]{vaswani_attention_2017}
Ashish Vaswani, Noam Shazeer, Niki Parmar, Jakob Uszkoreit, Llion Jones, Aidan~N Gomez, Łukasz Kaiser, and Illia Polosukhin.
\newblock Attention is {All} you {Need}.
\newblock In \emph{Advances in {Neural} {Information} {Processing} {Systems}}, volume~30. Curran Associates, Inc., 2017.

\bibitem[Vinyals et~al.(2019)Vinyals, Babuschkin, Czarnecki, Mathieu, Dudzik, Chung, Choi, Powell, Ewalds, Georgiev, and {others}]{vinyals_grandmaster_2019}
Oriol Vinyals, Igor Babuschkin, Wojciech~M Czarnecki, Michaël Mathieu, Andrew Dudzik, Junyoung Chung, David~H Choi, Richard Powell, Timo Ewalds, Petko Georgiev, and {others}.
\newblock Grandmaster {Level} in {StarCraft} {II} using {Multi}-{Agent} {Reinforcement} {Learning}.
\newblock \emph{Nature}, 575\penalty0 (7782):\penalty0 350--354, 2019.

\bibitem[Watter et~al.(2015)Watter, Springenberg, Boedecker, and Riedmiller]{watter_embed_2015}
Manuel Watter, Jost Springenberg, Joschka Boedecker, and Martin Riedmiller.
\newblock Embed to {Control}: {A} {Locally} {Linear} {Latent} {Dynamics} {Model} for {Control} from {Raw} {Images}.
\newblock In \emph{Advances in {Neural} {Information} {Processing} {Systems}}, volume~28. Curran Associates, Inc., 2015.

\bibitem[Wierstra et~al.(2007)Wierstra, Foerster, Peters, and Schmidhuber]{wierstra_solving_2007}
Daan Wierstra, Alexander Foerster, Jan Peters, and Jürgen Schmidhuber.
\newblock Solving {Deep} {Memory} {POMDPs} with {Recurrent} {Policy} {Gradients}.
\newblock In \emph{Artificial {Neural} {Networks}}, pages 697--706, 2007. Springer Berlin Heidelberg.

\bibitem[Zhang and Sennrich(2019)]{zhang_root_2019}
Biao Zhang and Rico Sennrich.
\newblock Root {Mean} {Square} {Layer} {Normalization}.
\newblock In \emph{Advances in {Neural} {Information} {Processing} {Systems}}, volume~32. Curran Associates, Inc., 2019.

\bibitem[Zhu et~al.(2017)Zhu, Mottaghi, Kolve, Lim, Gupta, Fei-Fei, and Farhadi]{zhu_target-driven_2017}
Yuke Zhu, Roozbeh Mottaghi, Eric Kolve, Joseph~J. Lim, Abhinav Gupta, Li~Fei-Fei, and Ali Farhadi.
\newblock Target-driven {Visual} {Navigation} in {Indoor} {Scenes} using {Deep} {Reinforcement} {Learning}.
\newblock In \emph{International {Conference} on {Robotics} and {Automation}}, 2017.

\bibitem[Zintgraf et~al.(2021)Zintgraf, Schulze, Lu, Feng, Igl, Shiarlis, Gal, Hofmann, and Whiteson]{zintgraf_varibad_2021}
Luisa Zintgraf, Sebastian Schulze, Cong Lu, Leo Feng, Maximilian Igl, Kyriacos Shiarlis, Yarin Gal, Katja Hofmann, and Shimon Whiteson.
\newblock {VariBAD}: {Variational} {Bayes}-{Adaptive} {Deep} {RL} via {Meta}-{Learning}.
\newblock \emph{Journal of Machine Learning Research}, 22\penalty0 (289):\penalty0 1--39, 2021.

\end{thebibliography}
\bibliographystyle{plainnatcustom}

\pagebreak
\appendix
\section{Associativity of Masked Associative Operators}
\label{app:associative_mao}
Let $\tilde{a}, \tilde{b}, \tilde{c} \in \tilde{\mathcal{E}}$ refer to elements in the
space of the MAO $\tilde{\bigcdot}$, as in \cref{def:mbo}, with $\tilde{a} = (a, m_a)$,
$\tilde{b} = (b, m_b)$, $\tilde{c} = (c, m_c)$. We show that if the sequence $\set{m_a,
m_b, m_c}$, is a right-padding mask, that is: $m_a = 1 \implies m_b = m_c = 1$, and $m_b
= 1 \implies m_c = 1$, then it holds that $(\tilde{a} \tilde{\bigcdot} \tilde{b})
\tilde{\bigcdot} \tilde{c} = \tilde{a} \tilde{\bigcdot} (\tilde{b} \tilde{\bigcdot}
\tilde{c})$, i.e., the MAO is associative. Similar to the proof in
\citet{lu_structured_2023} we consider all possible values for $\set{m_a,
m_b, m_c}$.

\paragraph{Case 1: $m_b = 1$ and $m_c = 1$.}
The binary masks of $b$ and $c$ are on, so $\tilde{b} \tilde{\bigcdot} \tilde{c} =
\tilde{b}$, $\tilde{a} \tilde{\bigcdot} \tilde{b} = \tilde{a}$ and $\tilde{a}
\tilde{\bigcdot} \tilde{c} = \tilde{a}$. Then,
\begin{align}
  (\tilde{a} \tilde{\bigcdot} \tilde{b}) \tilde{\bigcdot} \tilde{c} &= \tilde{a} \\
  &= \tilde{a} \tilde{\bigcdot} (\tilde{b} \tilde{\bigcdot} \tilde{c})
\end{align}
\paragraph{Case 2: $m_b = 0$ and $m_c = 1$.}
The binary mask of $b$ if off while that of $c$ is on, so $\tilde{b} \tilde{\bigcdot}
\tilde{c} = \tilde{b}$, then:
\begin{align}
  (\tilde{a} \tilde{\bigcdot} \tilde{b}) \tilde{\bigcdot} \tilde{c} &= \tilde{a} \tilde{\bigcdot} \tilde{b} \\
  &= \tilde{a} \tilde{\bigcdot} (\tilde{b} \tilde{\bigcdot} \tilde{c})
\end{align}
\paragraph{Case 3: $m_b = 0$ and $m_c = 0$.}
No mask is applied, then the MAO is equivalent to the underlying operator $\bigcdot$,
which is associative by \cref{def:mbo}.

Note the case $m_b = 1$ and $m_c = 0$ violates associativity, but it is impossible under
our initial assumption of a right-padding mask sequence $\set{m_a, m_b, m_c}$.

\section{Implementation Details}
\label{app:details}
In this section we provide details of various components of the RAC architecture and the
specific implementations of history encoders. All methods are implemented in a common
codebase written in the Pytorch framework \citep{paszke_pytorch_2019}.

\paragraph{Embedder.} We embed the concatenated observation-action history with a simple
linear layer mapping from the combined observation-action dimension to the embedding
dimension $E$.

\paragraph{Soft Actor-Critic.} We use a standard SAC implementation with optional
automatic entropy tuning \citep{haarnoja_soft_2019}. For discrete action spaces, we use
the discrete version of SAC by \citep{christodoulou_soft_2019} and one-hot encode the
actions.

\paragraph{\vanillassm, \ourmethod \& \noinput.} These methods share a similar
implementation, with an input linear layer, a linear recurrence and an output linear
layer. \vanillassm is equivalent to only using the ``Predict'' block from the KF layer,
while \noinput removes the input signal $u_{:t}$. For all methods, we discretize the SSM
using the zero-order hold method and a learnable scalar step size $\Delta$. In practice
we use an auxiliary learnable parameter $\tilde{\Delta}$ and define $\Delta =
\texttt{softplus}(\tilde{\Delta})$ to ensure a positive step size. as similarly done in
Mamba. We initialize $\tilde{\Delta}$ with a negative value such that after passing
through the softplus and after ZOH discretization, the SSM is initialized with
eigenvalues close to 1 (i.e., slow decay of state information over time).

\paragraph{\mamba.} Standard Mamba model from \citet{gu_mamba_2023}. We use a reference
open-source
implementation\footnote{\url{https://github.com/johnma2006/mamba-minimal/tree/03de542a36d873f6e6c4057ad687278cc6ae944d}}
and modify the parallel scan to use the associated MAO.

\paragraph{\gru.} Standard implementation included in Pytorch.

\paragraph{\transformer.} Default implementation of a causal transformer encoder from
Pytorch. We additionally include a sinusoidal positional encoding, as done in prior work
using transformers for RL \citep{ni_when_2023}.

\section{Hyperparameters}
\label{app:hparams}
\renewcommand{\arraystretch}{1.1}
\begin{table}[H]
\caption{Hyperparameters used for \cref{sec:experiments}. For the \mamba parameters, we
use the notation from the code by \citet{gu_mamba_2023} and select parameters to match a
effective state size $N = 128$. \gru and \transformer use default parameters from
Pytorch unless noted otherwise.}
\label{tab:hparam}
\begin{center}
\begin{tabular}{|c|c|c|c|}
\toprule
\textbf{Parameter}  & \textbf{BestArm} & \textbf{DMC} & \textbf{POPGym}\\
\midrule
\multicolumn{4}{|c|}{\textbf{Training}} \\
\midrule
Buffer size & \multicolumn{3}{c|}{$\infty$}\\
Adam learning rate & \multicolumn{3}{c|}{3e-4}\\
Env. steps & 500K & \multicolumn{2}{c|}{1M}\\
Batch size & 64 & \multicolumn{2}{c|}{32}\\
Update-to-data (UTD) ratio & 0.25 & \multicolumn{2}{c|}{$1.0$}\\
\# Eval episodes & $100$ & \multicolumn{2}{c|}{$16$}\\
\midrule
\multicolumn{4}{|c|}{\textbf{RAC}} \\
\midrule
Embedding size (E) & \multicolumn{3}{c|}{16}\\
Latent size (N) & \multicolumn{3}{c|}{128}\\
Activations & \multicolumn{3}{c|}{ReLU}\\
Context length  & 256 & \multicolumn{2}{c|}{64}\\
Actor MLP & [128] & \multicolumn{2}{c|}{[256, 256]}\\
Critic MLP & [256] & \multicolumn{2}{c|}{[256, 256]}\\
\midrule
\multicolumn{4}{|c|}{\textbf{SAC}} \\
\midrule
Discount factor $\gamma$ & \multicolumn{3}{c|}{0.99}\\
Entropy temp. $\alpha$ & 0.1 & \multicolumn{2}{c|}{Auto}\\
Target entropy (continuous) & N/A & \multicolumn{2}{c|}{-dim($\mathcal{A}$)}\\
Target entropy (discrete) & \multicolumn{2}{c|}{N/A} & $-0.7\log (1 / \text{dim}(\mathcal{A}))$\footnotemark \\
\midrule
\multicolumn{4}{|c|}{\textbf{History Encoders (common)}} \\
\midrule
Latent size $N$ & \multicolumn{3}{c|}{128}\\
\# layers & \multicolumn{3}{c|}{1}\\
\midrule
\multicolumn{4}{|c|}{\textbf{\vanillassm, \ourmethod \& \noinput}} \\
\midrule
$\tilde{\Delta}$ init & \multicolumn{3}{c|}{-7}\\
$\mat{A}$ init & \multicolumn{3}{c|}{HiPPO (diagonal)}\\
$\mat{B}$ init & \multicolumn{3}{c|}{$\mat{I}$}\\
$\mat{\Sigma}^{\textnormal{p}}$ init & \multicolumn{3}{c|}{$\mat{I}$}\\
Inital state belief & \multicolumn{3}{c|}{$\mathcal{N}(\mat{0}, \mat{I})$}\\
RMSNorm output? & No & Yes & No\\
\midrule
\multicolumn{4}{|c|}{\textbf{\mamba}} \\
\midrule
$\mat{A}$ init & \multicolumn{3}{c|}{HiPPO (diagonal)}\\
\texttt{d\textunderscore model} (embedding size) & \multicolumn{3}{c|}{16}\\
\texttt{d\textunderscore state} (per-channel hidden size) & \multicolumn{3}{c|}{4}\\
Expand factor $E$ & \multicolumn{3}{c|}{2}\\
Size of $\Delta$ projection & \multicolumn{3}{c|}{1}\\
1D Conv kernel size & \multicolumn{3}{c|}{4}\\
\midrule
\multicolumn{4}{|c|}{\textbf{\transformer}} \\
\midrule
\# heads & \multicolumn{3}{c|}{1}\\
Feedforward size & 128 & \multicolumn{2}{c|}{256}\\
\bottomrule
\end{tabular}
\end{center}
\end{table}
\footnotetext{We use a lower value of $-0.35\log (1 / \text{dim}(\mathcal{A}))$ in the
\texttt{MineSweeper} environment from POPGym, as the default value resulted in
divergence during training.}
\pagebreak

\section{Best Arm Identification Training Curves}
\label{sec:app_bestarm}

\begin{figure}[H]
  \centering
  \includegraphics[width=0.9\textwidth]{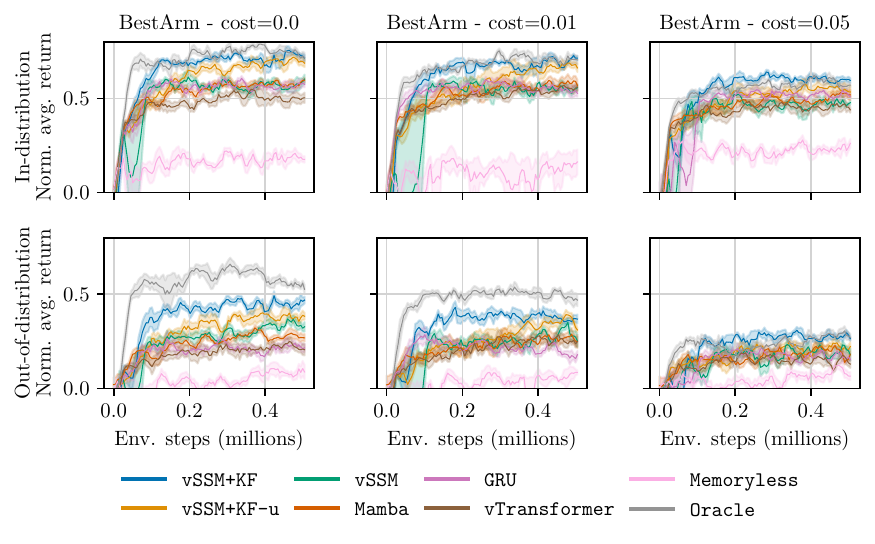}
  \caption{Normalized average return over 100 episodes in and out of distribution, for increasing costs. We report the mean and standard error over 5 random seeds.}
  \label{fig:bestarm_reward}
\end{figure}

\begin{figure}[H]
  \centering
  \includegraphics[width=0.9\textwidth]{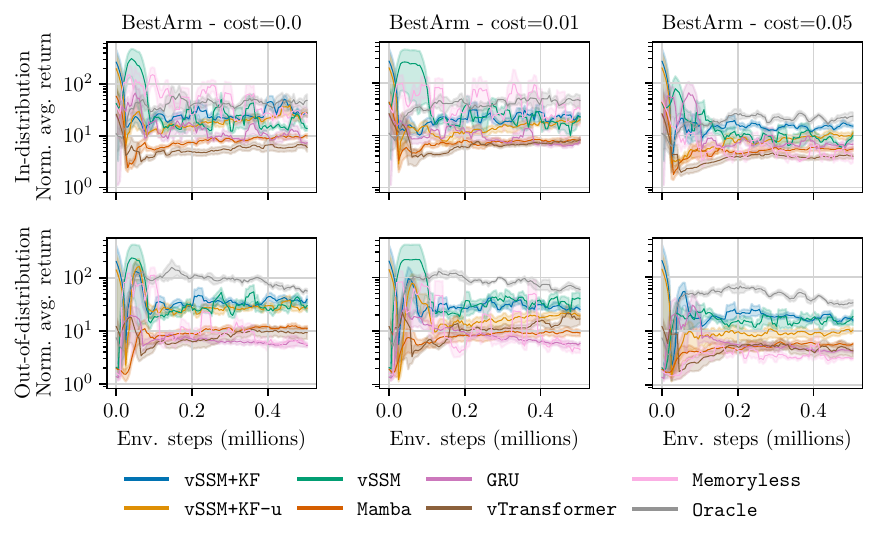}
  \caption{Average (log) episode length over 100 episodes in and out of distribution, for increasing costs. We report the mean and standard error over 5 random seeds.}
  \label{fig:bestarm_length}
\end{figure}

\section{KF Layer Design Ablation}
\label{sec:app_design_ablation}
\begin{figure}[H]
  \centering
  \includegraphics[width=0.9\textwidth]{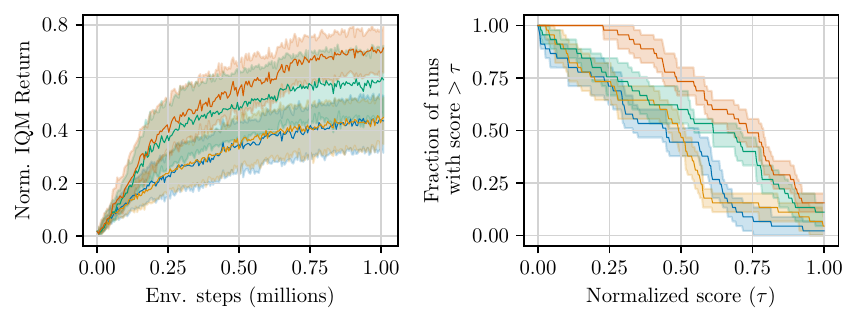}
  \includegraphics[width=0.9\textwidth]{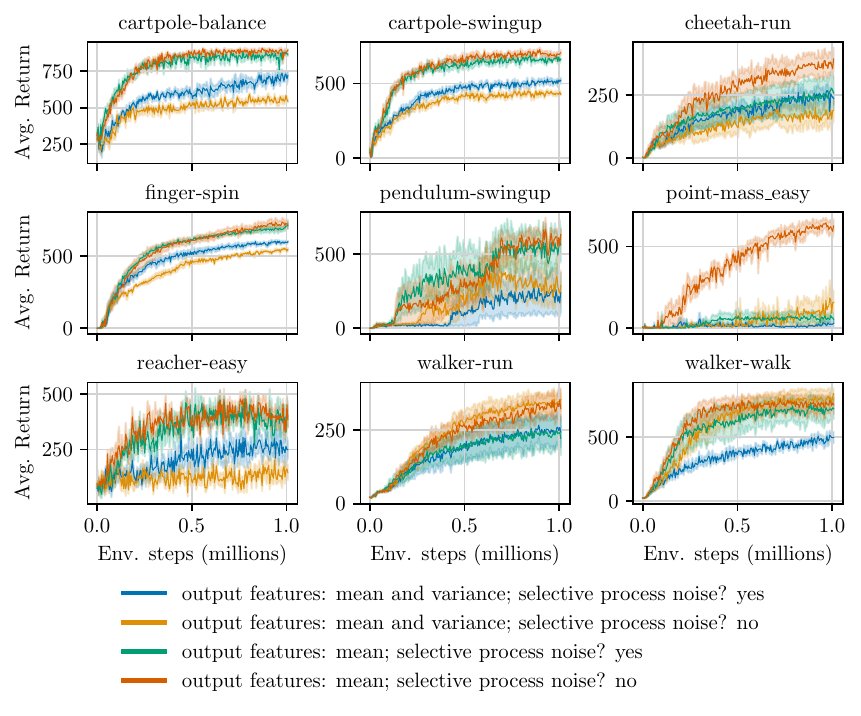}
  \caption{Ablation on design considerations for KF layers. \textbf{(Top)} Aggregated
  performance in noisy DMC benchmark (9 tasks) with 95\% bootstrap confidence intervals
  over five random seeds. \textbf{(Top-Left)} Inter-quartile mean returns normalized by
  the score of \oracle. \textbf{(Top-Right)} Performance profile after $1\textnormal{M}$
  environment steps. \textbf{(Bottom)} Training curves. We show mean and standard error
  over five random seeds. Based on these results, our final design for the KF layer uses
  only the posterior mean state as the output feature and a time-invariant process
  noise.}
  \label{fig:dmc_rand_design_ablation}
\end{figure}

\section{DMC Training Curves}
\label{sec:app_dmc_rand}

\begin{figure}[H]
  \centering
  \includegraphics[width=\textwidth]{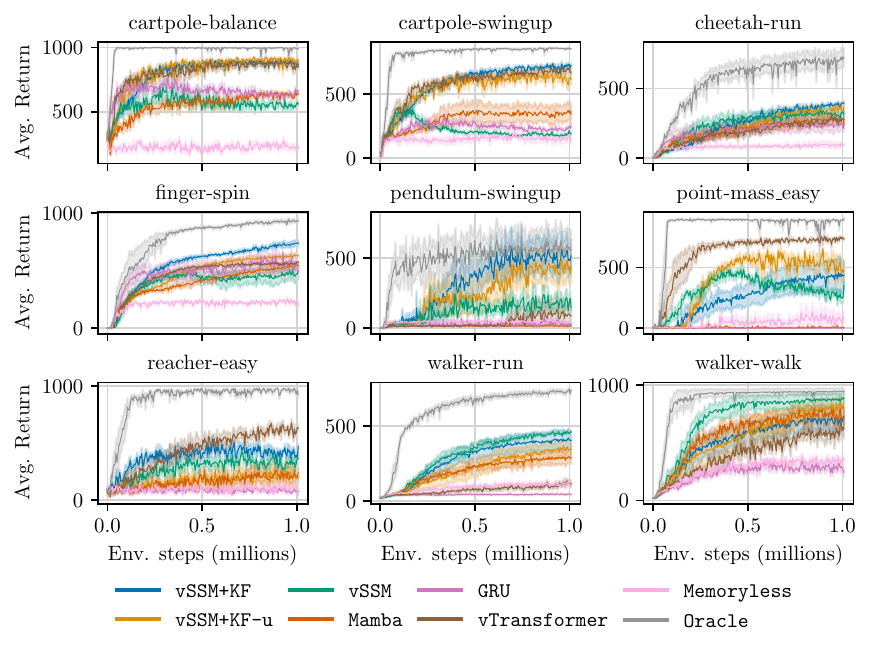}
  \caption{Training curves for the noisy DMC benchmark. We show mean and standard error
  over five random seeds. For all tasks, we add zero-mean Gaussian noise to the
  observations with a scale of $0.3$, except the \texttt{pendulum-swingup} and
  \texttt{point-mass} where the scale is $0.1$.}
  \label{fig:dmc_rand_full}
\end{figure}

\section{DMC Noise Ablation}
\label{sec:app_dmc_noise_ablation}

\begin{figure}[H]
  \centering
  \includegraphics[width=0.9\textwidth]{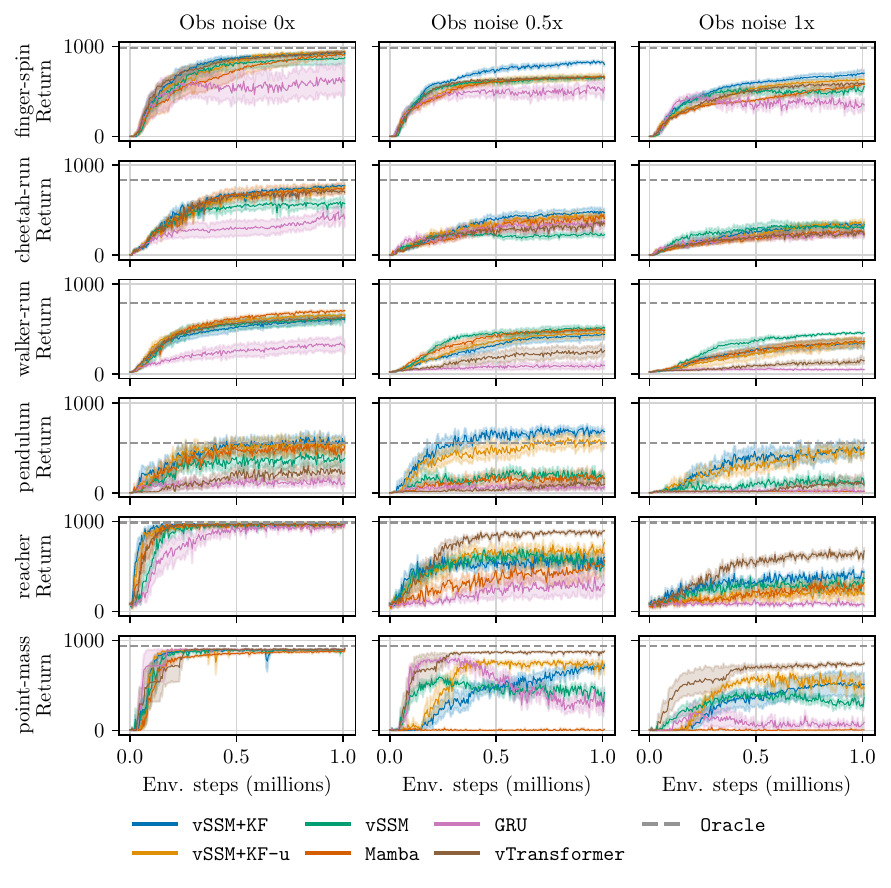}
  \caption{Training curves in six environments from the DMC benchmark with increasing
  levels of noise. We show mean and standard error over five random seeds (ten for
  \texttt{pendulum}). The base noise scale for all tasks is $0.3$, except the
  \texttt{pendulum-swingup} and \texttt{point-mass} environments where the scale is
  $0.1$}
  \label{fig:dmc_noise_ablation_full}
\end{figure}

\section{DMC Comparison to Model-Based Approaches}
\label{sec:app_dmc_modelbased}
\begin{figure}[H]
  \centering
  \includegraphics[width=0.9\textwidth]{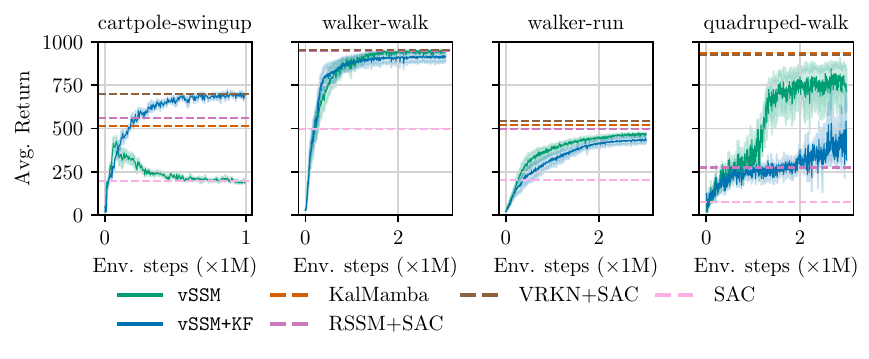}
  \caption{Comparison of performance in DMC environments with observation and
  action noise. For the model-based baselines, we report the final performance after 1M
  environment steps, as reported in \citet{becker_kalmamba_2024}.}
  \label{fig:dmc_noise_modelbased}
\end{figure}

In \cref{fig:dmc_noise_modelbased}, we compare performance between \vanillassm,
\ourmethod and the following model-based baselines reported in
\citet{becker_kalmamba_2024}\footnote{The experimental data was provided by the authors
on personal communication.}:

\paragraph{Kalmamba.} Uses a Mamba \citep{gu_mamba_2023} backbone that outputs the SSM
matrices, which are then used within the VRKN architecture proposed in
\citet{becker_uncertainty_2022}. The representation is trained end-to-end in a
variational loss to produce plausible dynamic predictions. The learned representation is
then frozen and used within a SAC policy optimizer.

\paragraph{RSSM+SAC.} The Recurrent SSM as proposed in \citet{hafner_learning_2019}.
Similarly, the representation is trained on a variational loss for reconstruction of
observations, while the policy is trained with SAC using the learned latent
representation. The gradients from SAC are not used to update the representation.

\paragraph{VRKN+SAC} The Variational Recurrent Kalman Network
\citep{becker_uncertainty_2022}. Similarly to RSSM+SAC and Kalmamba, the representation
is trained with a variational loss for reconstruction while the policy uses the learned
representation for control.

\paragraph{SAC.} Equivalent to the \nomem baseline; a SAC agent with no memory. 

The benchmark includes both observation \emph{and} action noise with a standard
deviation of $\sigma=0.3$.

From \cref{fig:dmc_noise_modelbased} we observe that in three out of four tasks,
\ourmethod is close to or matches the asymptotic performance of the best model-based
baseline, albeit with less sample-efficiency. These results highlight that good
performance can be achieved in these tasks without the representation learning
objectives from model-based approaches. In \texttt{quadruped-walk} we found that
probabilistic filtering hurts performance, given the performance difference between
\ourmethod and \vanillassm. We hypothesize that to handle the larger observation space
in \texttt{quadruped-walk} ($\sim4\times$ larger than \texttt{walker-run}), \ourmethod
would require further hyperparameter tuning and potential regularization techniques we
do not explore in this work. Moreover, the sample-efficiency of the architecture could
be further improved using recent developments such as higher update-to-data ratios and
network resets \citep{doro_sample-efficient_2022}.

\section{POPGym Training Curves}
\label{sec:app_popgym}

\begin{figure}[H]
  \centering
  \includegraphics[width=\textwidth]{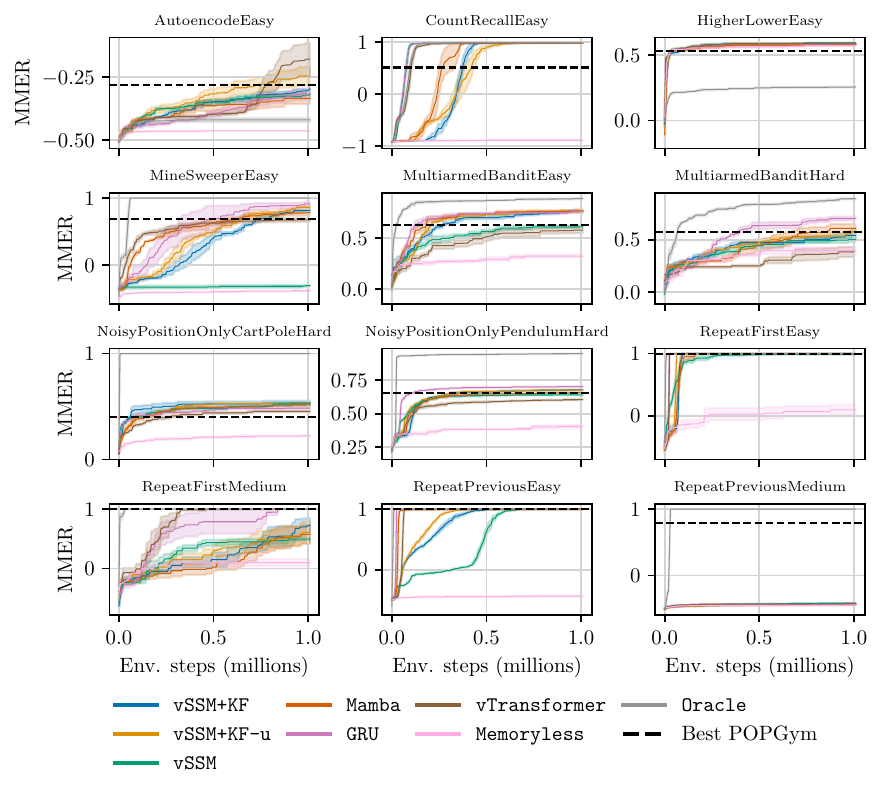}
  \caption{POPGym training curves. We show mean and standard error over five random seeds.}
  \label{fig:popgym_training_curves}
\end{figure}

\section{POPGym Scores}
\begin{figure}[H]
  \centering
  \includegraphics[width=\textwidth]{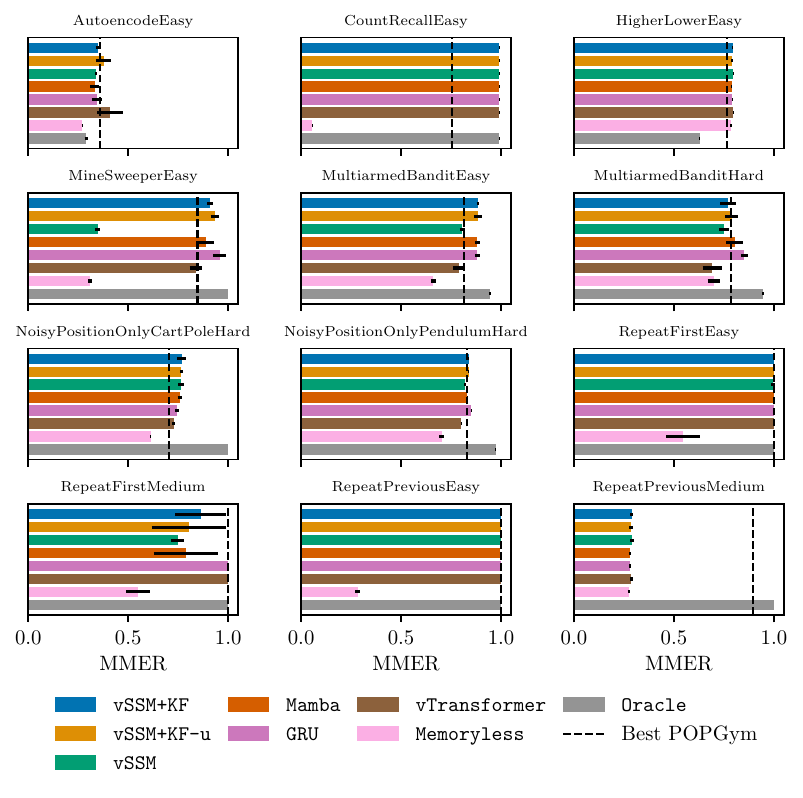}
  \caption{POPGym final MMER after 1M training steps. We show mean and standard error over five random seeds.}
  \label{fig:popgym_final_mmer}
\end{figure}

\renewcommand{\arraystretch}{1.5}
\begin{table}[!ht]
\caption{Scores on POPGym tasks after 1M environment steps. For each environment, we
report the MMER mean and standard error over 5 random seeds after 1M steps of training.
The MMER is calculated from 16 test episodes. For reference, we  include the best MMER
score reported by \citet{morad_popgym_2023} (mean and standard deviation over three
random seeds). We \textbf{bold} the highest score(s) per environment obtained by a
sequence model.}
\label{tab:dm_control_scores}
\centerline{
\begin{tabular}{|c|cccc|}
\cline{2-5}
\multicolumn{1}{c|}{} & AutoencodeEasy & CountRecallEasy & HigherLowerEasy & MineSweeperEasy \\
\midrule
\ourmethod & $-0.299 \pm 0.012$ & $\mathbf{0.983} \pm 0.002$ & $0.588 \pm 0.002$ & $0.818 \pm 0.014$ \\
\noinput & $-0.247 \pm 0.036$ & $\mathbf{0.978} \pm 0.001$ & $0.581 \pm 0.004$ & $0.864 \pm 0.020$ \\
\vanillassm & $-0.320 \pm 0.006$ & $\mathbf{0.984} \pm 0.002$ & $\mathbf{0.592} \pm 0.003$ &
$-0.307 \pm 0.012$ \\
\mamba & $-0.335 \pm 0.023$ & $0.982 \pm 0.002$ & $0.580 \pm 0.003$ & $0.783 \pm 0.039$ \\
\gru & $-0.317 \pm 0.025$ & $\mathbf{0.984} \pm 0.001$ & $0.586 \pm 0.002$ &
$\mathbf{0.916} \pm 0.034$ \\
\transformer & $\mathbf{-0.179} \pm 0.065$ & $0.982 \pm 0.001$ & $\mathbf{0.590} \pm 0.001$ & $0.676 \pm 0.031$ \\
\midrule
\oracle & $-0.420 \pm 0.008$ & $0.984 \pm 0.002$ & $0.257 \pm 0.003$ & $1.000 \pm 0.000$
\\
\nomem & $-0.463 \pm 0.001$ & $-0.887 \pm 0.001$ & $0.571 \pm 0.004$ & $-0.382 \pm 0.008$ \\
Best POPGym & $-0.283\pm0.029$ & $0.509\pm0.062$ & $0.529\pm0.002$ & $0.693\pm0.009$ \\
\bottomrule
\end{tabular}
}
\vspace{1em}
\centerline{
\begin{tabular}{|c|cccc|}
\cline{2-5}
\multicolumn{1}{c|}{} & BanditEasy & BanditHard & NoisyCartPoleHard & NoisyPendulumHard \\
\midrule
\ourmethod & $\mathbf{0.766} \pm 0.005$ & $0.541 \pm 0.040$ & $\mathbf{0.535} \pm 0.023$ & $0.677
\pm 0.003$ \\
\noinput & $\mathbf{0.771} \pm 0.019$ & $0.579 \pm 0.032$ & $\mathbf{0.531} \pm 0.007$ & $0.675
\pm 0.003$ \\
\vanillassm & $0.612 \pm 0.013$ & $0.501 \pm 0.025$ & $0.528 \pm 0.013$ & $0.639 \pm 0.004$ \\
\mamba & $\mathbf{0.764} \pm 0.011$ & $0.608 \pm 0.043$ & $\mathbf{0.516} \pm 0.010$ & $0.658 \pm 0.009$ \\
\gru & $\mathbf{0.763} \pm 0.012$ & $\mathbf{0.705} \pm 0.017$ & $0.486 \pm 0.010$ &
$\mathbf{0.701} \pm 0.001$ \\
\transformer & $0.580 \pm 0.030$ & $0.384 \pm 0.049$ & $0.454 \pm 0.009$ & $0.604 \pm 0.004$ \\
\midrule
\oracle & $0.889 \pm 0.005$ & $0.892 \pm 0.006$ & $1.000 \pm 0.000$ & $0.946 \pm 0.001$ \\
\nomem & $0.324 \pm 0.013$ & $0.399 \pm 0.031$ & $0.225 \pm 0.003$ & $0.406 \pm 0.012$ \\
Best POPGym & $0.631\pm0.014$ & $0.574\pm0.049$ & $0.404\pm0.005$ & $0.657\pm0.002$ \\
\bottomrule
\end{tabular}
}
\vspace{1em}
\centerline{
\begin{tabular}{|c|cccc|}
\cline{2-5}
\multicolumn{1}{c|}{} & RepeatFirstEasy & RepeatFirstMedium & RepeatPreviousEasy &
RepeatPreviousMedium \\
\midrule
\ourmethod & $\mathbf{1.000} \pm 0.000$ & $0.726 \pm 0.127$ & $\mathbf{1.000} \pm 0.000$ & $\mathbf{-0.423} \pm 0.006$ \\
\noinput & $\mathbf{1.000} \pm 0.000$ & $0.607 \pm 0.186$ & $\mathbf{1.000} \pm 0.000$ & $\mathbf{-0.429} \pm 0.008$ \\
\vanillassm & $0.989 \pm 0.009$ & $0.495 \pm 0.034$ & $\mathbf{1.000} \pm 0.000$ &
$\mathbf{-0.420} \pm 0.008$ \\
\mamba & $\mathbf{1.000} \pm 0.000$ & $0.575 \pm 0.160$ & $0.993 \pm 0.001$ & $-0.441 \pm 0.005$ \\
\gru & $\mathbf{1.000} \pm 0.000$ & $\mathbf{1.000} \pm 0.000$ & $\mathbf{1.000} \pm 0.000$ & $-0.440 \pm 0.004$ \\
\transformer & $\mathbf{1.000} \pm 0.000$ & $\mathbf{1.000} \pm 0.000$ & $\mathbf{1.000} \pm 0.000$ & $\mathbf{-0.426} \pm 0.006$ \\
\midrule
\oracle & $1.000 \pm 0.000$ & $1.000 \pm 0.000$ & $1.000 \pm 0.000$ & $1.000 \pm 0.000$
\\
\nomem & $0.093 \pm 0.085$ & $0.100 \pm 0.061$ & $-0.434 \pm 0.013$ & $-0.450 \pm 0.007$ \\
Best POPGym & $1.000\pm0.000$ & $1.000\pm0.000$ & $1.000\pm0.000$ & $0.789\pm0.288$ \\
\bottomrule
\end{tabular}
}
\end{table}

\section{POPGym Ablation}
\label{app:popgym_ablation}
\begin{figure}[H]
  \centering
  \includegraphics[width=\textwidth]{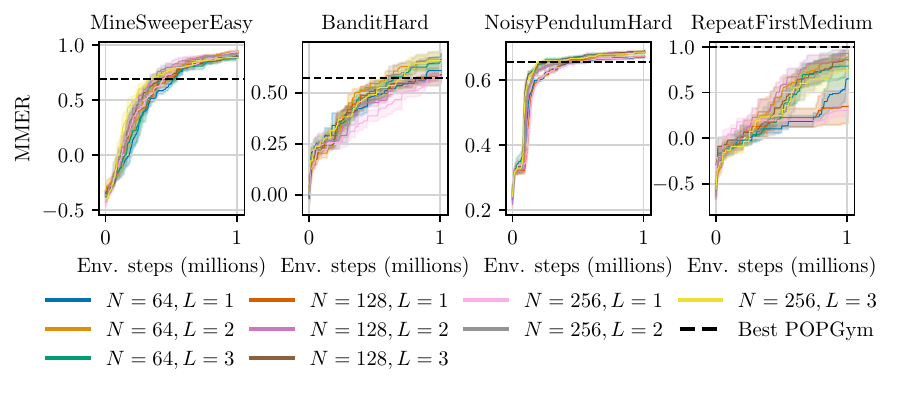}
  \caption{POPGym training curves for \ourmethod ablation over latent state size $N$ and number of KF layers $L$. We show mean and standard error over five random seeds. For this experiment, \ourmethod uses an RMSNorm output block to ensure stability for $L>1$.}
  \label{fig:popgym_ablation_full}
\end{figure}

\end{document}